\documentclass[twocolumn,journal]{IEEEtran}
\usepackage[T1]{fontenc}
\usepackage[latin9]{inputenc}
\usepackage{array}
\usepackage{float}
\usepackage{bm}
\usepackage{multirow}
\usepackage{amsmath}
\usepackage{amsthm}
\usepackage{amssymb}
\usepackage{graphicx}
\usepackage[unicode=true,
 bookmarks=true,bookmarksnumbered=true,bookmarksopen=true,bookmarksopenlevel=1,
 breaklinks=false,pdfborder={0 0 0},pdfborderstyle={},backref=false,colorlinks=false]
 {hyperref}
\hypersetup{pdftitle={Your Title},
 pdfauthor={Your Name},
 pdfpagelayout=OneColumn, pdfnewwindow=true, pdfstartview=XYZ, plainpages=false}

\makeatletter

\providecommand{\tabularnewline}{\\}
\floatstyle{ruled}
\newfloat{algorithm}{tbp}{loa}
\providecommand{\algorithmname}{Algorithm}
\floatname{algorithm}{\protect\algorithmname}

\let\oldforeign@language\foreign@language
\DeclareRobustCommand{\foreign@language}[1]{%
  \lowercase{\oldforeign@language{#1}}}
\theoremstyle{plain}
\newtheorem{thm}{\protect\theoremname}
\theoremstyle{plain}
\newtheorem{lem}[thm]{\protect\lemmaname}

\usepackage[caption=false,font=footnotesize]{subfig}

\usepackage{cite}

\allowdisplaybreaks

\makeatletter
\newcommand{\vast}{\bBigg@{4}}
\newcommand{\Vast}{\bBigg@{5}}
\makeatother



\providecommand{\lemmaname}{Lemma}
\providecommand{\theoremname}{Theorem}

\begin{document}
\title{Decentralized Complete Dictionary Learning via $\ell^{4}$-Norm Maximization}
\author{Qiheng~Lu,~Lixiang Lian\thanks{Qiheng~Lu and Lixiang Lian are with the School of Information Science
and Technology, ShanghaiTech University, Shanghai, China, e-mail:
\protect\href{mailto:\%7Bluqh,\%20lianlx\%7D@shanghaitech.edu.cn}{\{luqh, lianlx\}@shanghaitech.edu.cn}.}}
\markboth{Journal of XXX}{Qiheng Lu \MakeLowercase{\emph{et al.}}: Decentralized Complete
Dictionary Learning via $\ell^{4}$-Norm Maximization}
\maketitle
\begin{abstract}
With the rapid development of information technologies, centralized
data processing is subject to many limitations, such as computational
overheads, communication delays, and data privacy leakage. Decentralized
data processing over networked terminal nodes becomes an important
technology in the era of big data. Dictionary learning is a powerful
representation learning method to exploit the low-dimensional structure
from the high-dimensional data. By exploiting the low-dimensional
structure, the storage and the processing overhead of data can be
effectively reduced. In this paper, we propose a novel decentralized
complete dictionary learning algorithm, which is based on $\ell^{4}$-norm
maximization. Compared with existing decentralized dictionary learning
algorithms, comprehensive numerical experiments show that the novel
algorithm has significant advantages in terms of per-iteration computational
complexity, communication cost, and convergence rate in many scenarios.
Moreover, a rigorous theoretical analysis shows that the dictionaries
learned by the proposed algorithm can converge to the one learned
by a centralized dictionary learning algorithm at a linear rate with
high probability under certain conditions.
\end{abstract}


\IEEEpeerreviewmaketitle{}

\section{Introduction}

\IEEEPARstart{W}{ith} the development of the Internet of Things
(IoT), sensors and various data acquisition devices are ubiquitous
and widely used in intelligent transportation, smart grids, and other
fields. The centralized data processing method requires network terminal
nodes to transmit locally generated data to a central node for data
analysis and processing. However, transmitting a large amount of local
data can cause network congestion, communication delay, data privacy
leakage, and other problems, which limit the real-time decision in
the network. Moreover, since the centralized node needs to process
all the data, it has too much computational overhead, which limits
the application of this processing method in large-scale scenarios.
Since a lot of data is generated at network terminals, decentralized
real-time data processing at network terminal nodes becomes an important
technology to solve the above problems during data transmission.

In the era of big data, it has been observed that many high-dimensional
real-world data present a low-dimensional structure under the projection
of proper bases \cite{wright2022high}. The low-dimensional structure
of big data can be effectively exploited to complete various signal
processing tasks with reduced resource consumption, such as image
denoising \cite{elad2006image}, face recognition \cite{wright2009robust},
and matrix completion \cite{candes2009exact}. Therefore, it is critical
to learn the low-dimensional structure underlying big data. Dictionary
learning (DL) is a representation learning method that can efficiently
find low-dimensional structures in high-dimensional data. Given the
observed data $\bm{Y}$, DL aims to decompose the observation matrix
$\bm{Y}$ into a dictionary matrix $\bm{D}$ and a sparse coefficient
matrix $\bm{X}$ such that $\bm{Y}=\bm{DX}$ or $\bm{Y}\approx\bm{DX}$.
The concept of DL was firstly introduced by Olshausen and Field \cite{olshausen1996emergence,olshausen1997sparse},
after which, many DL algorithms have been proposed including \textit{Method
of Optimal Directions} (MOD) \cite{engan2000multi}, K-SVD \cite{aharon2006k},
and \textit{Matching, Stretching, and Projection} (MSP) \cite{zhai2020complete}.
However, these algorithms are centralized algorithms, which require
the collection of data over the network and thus cause overhead problems
or privacy issues.

In this paper, we focus on the problem of decentralized DL, where
the decentralized nodes collaboratively learn the dictionary from
the local data by exchanging information among neighboring nodes.
The decentralized DL shares the advantages of low latency, low transmission
overheads, and data privacy protection, and at the same time, the
compressibility of high-dimensional data is effectively utilized through
dictionary learning to efficiently process high-dimensional data.

\subsection{Previous Work}

In the past decades, several decentralized DL algorithms have been
proposed \cite{chainais2013learning,liang2014distributed,chouvardas2015an,wai2015a,chen2015dictionary,zhao2016a,raja2016cloud,li2019the,daneshmand2019decentralized,dong2022distributed,wang2022decentralized}.
Those algorithms can be divided into two main categories, the first
one is based on centralized DL algorithms and the second one is based
on decentralized frameworks. \cite{liang2014distributed,raja2016cloud,dong2022distributed}
belong to the first category, where \cite{liang2014distributed,raja2016cloud}
are based on K-SVD \cite{aharon2006k} and \cite{dong2022distributed}
is based on Analysis SimCO \cite{dong2016analysis}. However, these
algorithms are not attractive in terms of convergence rate, communication
cost, per-iteration computational complexity, etc. Specifically, the
per-iteration computational complexity of all three algorithms is
very high because they are all based on computational-intensive centralized
algorithms. Meanwhile, \cite{liang2014distributed,dong2022distributed}
need thousands of iterations to achieve a good convergence result,
thus introducing excessive communication overheads. \cite{chainais2013learning,chouvardas2015an,wai2015a,chen2015dictionary,zhao2016a,li2019the,daneshmand2019decentralized,wang2022decentralized}
belong to the second category, and they are based on some novel or
existing decentralized frameworks, such as EXTRA \cite{shi2015extra},
that can be applied directly or with minor modifications to DL. Similarly,
these algorithms are also not satisfactory in terms of convergence
performance, convergence rate, communication cost, etc. Specifically,
some algorithms cannot converge. For example, numerical experiments
in \cite{daneshmand2019decentralized} showed that the decentralized
algorithm proposed in \cite{chainais2013learning} failed to reach
consensus among local dictionaries. To the best of our knowledge,
the fastest convergence rate proven among these algorithms can only
achieve a sub-linear rate \cite{daneshmand2019decentralized,wang2022decentralized},
which leads to this category of algorithms all requiring thousands
of iterations to achieve a good convergence result. The difficulties
in improving the performance of these two categories of existing decentralized
DL algorithms are as follows:
\begin{itemize}
\item The key to proposing an effective and efficient decentralized DL algorithm
based on existing centralized DL algorithms is to find a centralized
algorithm that is effective, efficient and easy to be decentralized.
However, although many novel centralized DL formulations and algorithms,
such as \cite{schramm2017fast,sun2017complete1,bai2018subgradient},
proposed in recent years have elaborate designs and rigorous convergence
analyses, they are not efficient and effective enough in practice,
especially in large-scale scenarios.
\item Most of the existing decentralized frameworks and optimization algorithms
that can be applied to DL are based on $\ell^{1}$-norm penalization.
However, solving a non-convex optimization problem whose objective
function is non-smooth is NP-hard in general \cite{murty1987some},
which means that it is very difficult or even impossible to design
an efficient and effective decentralized DL algorithm based on such
formulation.
\end{itemize}
In the past few years, the development of centralized DL algorithms
can inspire us to design more advanced decentralized DL algorithms.
It was shown in \cite{sun2017complete1} that if the sparse matrix
$\bm{X}$ obeys the Bernoulli-Gaussian distribution, the complete
DL problem can be converted to an orthogonal complete DL problem without
loss of generality by preconditioning $\bm{Y}$. \cite{bao2013fast}
showed that the orthogonal complete DL algorithm proposed therein
can achieve comparable learning performance compared with the overcomplete
DL algorithms applied for image restoration, but with significantly
lower computational cost. \cite{zhai2020complete} proposed an orthogonal
complete DL algorithm MSP based on $\ell^{4}$-norm maximization.
Because of the benign global landscape of the objective function,
the proposed algorithm can easily converge to a global optimum. Another
attractive aspect of the $\ell^{4}$-based DL algorithm is its low
computational complexity and super-linear convergence rate.

Although the $\ell^{4}$-based formulation has appealing properties,
the non-convex objective function and the orthogonality constraint
of such formulation make these algorithms hard to be decentralized.
Decentralized constrained non-convex optimization algorithms have
been widely studied over the past decade or so, but no decentralized
algorithm is applicable to the orthogonality constraint until DRSGD
and DRGTA were proposed in 2021 \cite{chen2021decentralized}. The
gradient of the objective function of $\ell^{4}$-norm maximization
is not $L$-Lipschitz continuous, which does not meet the requirements
of DRSGD and DRGTA, so they are also not applicable to $\ell^{4}$-based
formulation. To the best of our knowledge, DESTINY \cite{wang2022decentralized},
proposed very recently, is the first decentralized framework that
can be applied to $\ell^{4}$-based formulation. However, DESTINY
can only achieve a sub-linear convergence rate, while MSP can achieve
a super-linear convergence rate. Therefore, DESTINY cannot fully utilize
the advantages of the $\ell^{4}$-based formulation. Besides, DESTINY
also requires the topology of the decentralized network to be a connected
undirected graph, a requirement that cannot be satisfied in some practical
scenarios.

\subsection{Our Contributions}

In this paper, we propose a decentralized complete DL algorithm based
on $\ell^{4}$-norm maximization, termed \textit{Decentralized Matching,
Stretching, and Projection} (DMSP), which is a decentralized variant
of MSP to learn the dictionaries only based on local data. The main
contributions of this paper are summarized as follows:
\begin{itemize}
\item We propose a decentralized DL algorithm DMSP that takes full advantage
of the $\ell^{4}$-based formulation, which gives DMSP  a good performance
in terms of the per-iteration computational complexity, communication
cost, convergence rate, etc.
\item We prove that the dictionaries learned by DMSP can converge to the
one learned by MSP at a linear rate with high probability under certain
conditions.
\item We present extensive numerical experiments to show that DMSP can converge
to MSP under rather mild conditions and by fully utilizing the advantages
of $\ell^{4}$-based formulation, the proposed DMSP achieves unmatched
performance in terms of the per-iteration computational complexity,
communication cost, convergence rate, compared with various baselines. 
\end{itemize}

\subsection*{Notations}

In this paper, lowercase regular letters denote scalars, lowercase
bold letters denote vectors, and uppercase bold letters denote matrices.
$x_{i}$ denotes the $i^{\text{th}}$ entry of vector $\bm{x}$, $\bm{x}_{i,\ast}$
denotes the $i^{\text{th}}$ row of matrix $\bm{X}$, $\bm{x}_{\ast,i}$
denotes the $i^{\text{th}}$ column of matrix $\bm{X}$, and $x_{i,j}$
denotes the entry in the $i^{\text{th}}$ row and $j^{\text{th}}$
column of matrix $\bm{X}$. $\bm{X}_{i}$ denotes a local matrix at
node $i$, $\bm{X}^{(t)}$ denotes a matrix at the $t^{\text{th}}$
iteration, and $\bm{X}^{(t,s)}$ denotes a matrix at the $s^{\text{th}}$
inner iteration in the $t^{\text{th}}$ outer iteration. $\Vert\cdot\Vert_{4}$
denotes the element-wise $\ell^{4}$-norm, $\Vert\cdot\Vert_{F}$
denotes the Frobenius norm. $[N]$ denotes a set $\{1,2,\ldots,N\}$.
$(\bm{A}_{i})_{i\in[N]}$ denotes a tuple $(\bm{A}_{1},\ldots,\bm{A}_{N})$.
$\circ$ denotes the Hadamard product and $(\cdot)^{\circ r}$ denotes
the $r^{\text{th}}$ Hadamard power. $\mathcal{P}_{\bm{G}}(\cdot)$
denotes a projection onto group $\bm{G}$. $\mathrm{O}(n)$ denotes
the $n$-dimensional orthogonal group.

\section{Problem Formulation }

Consider a decentralized network with $N$ nodes and this network
can be abstracted as a time-varying directed graph sequence $\left\{ \mathcal{G}^{(t,s)}\right\} _{t,s}$
, where $\mathcal{G}^{(t,s)}=\left(\mathcal{V},\mathcal{E}^{(t,s)}\right)$
is the graph at the $s^{\text{th}}$ inner iteration in the $t^{\text{th}}$
outer iteration, $\mathcal{V}\in[N]$, and $(i,j)\in\mathcal{E}^{(t,s)}$
if and only if a node $i$ can send messages to another node $j$
at the $s^{\text{th}}$ inner iteration in the $t^{\text{th}}$ outer
iteration. Define the in-neighborhood of node $i$ at the $s^{\text{th}}$
inner iteration in the $t^{\text{th}}$ outer iteration as $\mathcal{N}_{i,in}^{(t,s)}=\left\{ j\in\mathcal{V}|(j,i)\in\mathcal{E}^{(t,s)}\right\} \cup\{i\}$
and define the out-neighborhood of node $i$ at the $s^{\text{th}}$
inner iteration in the $t^{\text{th}}$ outer iteration as $\mathcal{N}_{i,out}^{(t,s)}=\left\{ j\in\mathcal{V}|(i,j)\in\mathcal{E}^{(t,s)}\right\} \cup\{i\}$.
We assume that the graph sequence $\left\{ \mathcal{G}^{(t,s)}\right\} _{t,s}$
is uniformly strongly connected, i.e., there exists a positive integer
$B$, such that $\mathcal{G}^{(t)}=\left(\mathcal{V},\bigcup_{i=kB}^{(k+1)B-1}\mathcal{E}^{(t,i)}\right)$
is strongly connected for every $k>0$ and every $t>0$. Let $\bm{W}^{(t,s)}\in\mathbb{R}^{n\times n}$
denote the weighted adjacency matrix at the $s^{\text{th}}$ inner
iteration in the $t^{\text{th}}$ outer iteration. The weighted adjacency
matrix $\bm{W}^{(t,s)}$ needs to satisfy the following assumptions:
$\bm{W}^{(t,s)}$ is a column stochastic matrix and $w_{i,j}^{(t,s)}>0$
if and only if $j\in\mathcal{N}_{i,in}^{(t,s)}$.

Suppose that each node $i$ generates its local observation matrix
$\bm{Y}_{i}\in\mathbb{R}^{n\times p_{i}}$ which can be expressed
as $\bm{Y}_{i}=\bm{D}_{o}\bm{X}_{i}$, where $\bm{D}_{o}\in\mathbb{R}^{n\times n}$
is the ground truth dictionary and $\bm{X}_{i}\in\mathbb{R}^{n\times p_{i}}$
is a Bernoulli-Gaussian random matrix with sparsity $\theta$, i.e.,
$\{\bm{X}_{i}\}_{j,k}\sim_{i.i.d.}\mathrm{BG}(\theta)$. Since the
complete DL problem can be converted to an orthogonal complete DL
problem without loss of generality under the Bernoulli-Gaussian model,
we consider orthogonal complete DL in this paper, which means $\bm{D}_{o}$
is a square matrix satisfying the orthogonality constraint.

Let $\bm{Y}=[\bm{Y}_{1},\ldots,\bm{Y}_{N}]\in\mathbb{R}^{n\times p}$
denote the collection of local data, where $p=\sum_{i\in[N]}p_{i}$.
If all local data are available at a fusion center, an $\ell^{4}$-based
orthogonal complete DL problem can be formulated as

\begin{equation}
\begin{aligned}\max_{\bm{D},\bm{X}} & \quad\Vert\bm{X}\Vert_{4}^{4}\\
\mathrm{s.t.} & \quad\bm{Y}=\bm{DX},\\
 & \quad\boldsymbol{D}\in\mathrm{O}(n).
\end{aligned}
\label{eq:eq1}
\end{equation}

Let $\bm{A}=\bm{D}^{\mathsf{T}}$, (\ref{eq:eq1}) can be reformulated
as

\begin{equation}
\begin{aligned}\max_{\bm{A}} & \quad\Vert\bm{AY}\Vert_{4}^{4}\\
\mathrm{s.t.} & \quad\bm{A}\in\mathrm{O}(n).
\end{aligned}
\label{eq:eq2}
\end{equation}

However, data collection in a centralized fusion center can induce
huge communication overheads and leak data privacy. In this paper,
we aim to design a decentralized orthogonal complete DL algorithm
based on $\ell^{4}$-norm maximization, so that the nodes in the network
can collaboratively learn the dictionary without exchanging the raw
data.

Suppose that $\bm{P}\in\mathbb{R}^{n\times n}$ is a signed permutation
matrix. Since $\bm{D}_{o}$ and $\bm{D}_{o}\bm{P}^{\mathsf{T}}$ are
both orthogonal, $\bm{X}$ and $\bm{PX}$ have the same sparsity level,
i.e., $\Vert\bm{X}\Vert_{0}=\Vert\bm{PX}\Vert_{0}$, the orthogonal
complete DL problem is signed permutation ambiguous. Therefore, it
is considered to be perfectly recovery if any signed permutation of
the ground truth dictionary is found.

\section{$\ell^{4}$-Based Orthogonal Complete Dictionary Learning Algorithm}

\subsection{Centralized $\ell^{4}$-Based Dictionary Learning Algorithm}

Most of existing DL algorithms are based on $\ell^{1}$-norm minimization
to promote sparsity. An $\ell^{1}$-based orthogonal complete DL problem
can be formulated as

\begin{equation}
\begin{aligned}\min_{\bm{A}} & \quad\Vert\bm{AY}\Vert_{1}\\
\mathrm{s.t.} & \quad\bm{A}\in\mathrm{O}(n).
\end{aligned}
\label{eq:eq3}
\end{equation}

However, due to the non-smoothness of $\ell^{1}$-norm and the non-convexity
of orthogonality constraint, (\ref{eq:eq3}) is NP-hard in general
\cite{murty1987some}. On the contrary, the smoothness of $\ell^{4}$-norm
and the sparsity imposition of $\ell^{4}$-norm maximization enable
a more efficient DL algorithm design.

Designing an efficient and effective optimization algorithm to solve
(\ref{eq:eq2}) is tough because of the non-convex objective function
and the orthogonality constraint of (\ref{eq:eq2}). Based on Algorithm
1 in \cite{journee2010generalized}, \cite{zhai2020complete} proposed
MSP (see Algorithm \ref{alg:alg1}) to solve (\ref{eq:eq2}). Each
iteration of MSP performs a projected gradient ascent with infinite
step size, and MSP is thus able to achieve a super-linear convergence
rate. Numerical experiments in \cite{zhai2020complete,zhai2020understanding}
verified that MSP outperforms existing DL algorithms in many scenarios
(including noise, outliers, and sparse corruptions).

\begin{algorithm}[tbh]
\caption{MSP (Algorithm 2 in \cite{zhai2020complete})\label{alg:alg1}}

1: \textbf{Initialize $\bm{A}^{(0)}\in\mathrm{O}(n)$}

2: \textbf{for $t=0,1,\ldots,T-1$ do}

3: $\quad\partial\bm{A}^{(t)}\gets4\left(\bm{A}^{(t)}\bm{Y}\right)^{\circ3}\bm{Y}^{\mathsf{T}}$

4: $\quad\bm{U\Sigma V}^{\mathsf{T}}\gets\mathrm{SVD}\left(\partial\bm{A}^{(t)}\right)$

5: $\quad\bm{A}^{(t+1)}\gets\bm{UV}^{\mathsf{T}}$

6: \textbf{end for}

7:\textbf{ Output $\bm{A}^{(T)}$}
\end{algorithm}

\subsection{Decentralized $\ell^{4}$-Based Dictionary Learning Algorithm}

The key to proposing an efficient and effective $\ell^{4}$-based
decentralized orthogonal complete DL algorithm is to find out which
part of MSP can be implemented in a decentralized manner. From observation,
Step 3 and Step 5 of MSP have the potential for decentralized framework
design.

If the decentralized variant of MSP is realized by designing the decentralized
framework for Step 5 of MSP, consensus averaging can be employed to
exchange the dictionaries learned at the individual node $\bm{A}_{1}^{(t)}$,
$\ldots$, $\bm{A}_{N}^{(t)}$ after each iteration. However, this
decentralized framework cannot guarantee to satisfy the orthogonality
constraint in (\ref{eq:eq2}). Specifically, the orthogonal group
is non-convex, so locally estimated dictionaries after consensus averaging
might not be orthogonal, which violates the orthogonality constraint
in (\ref{eq:eq2}). If each node projects the linearly combined estimated
matrix onto the orthogonal group again, the convergence of this decentralized
framework cannot be guaranteed even if the perfect consensus is achieved
in every iteration because projection onto a non-convex set is not
non-expansive.

This paper proposes the algorithm DMSP (see Algorithm \ref{alg:alg2})
by designing the decentralized framework for Step 3 in MSP. DMSP performs
consensus averaging to exchange the local gradient matrix of each
node $\partial\bm{A}_{1}^{(t)}$, $\ldots$, $\partial\bm{A}_{N}^{(t)}$
in each iteration. By exchanging the gradient instead of raw data,
the proposed algorithm DMSP can protect data privacy. This design
is effective because MSP has two useful properties:
\begin{itemize}
\item Separable: $(\bm{AY})^{\circ3}\bm{Y}^{\mathsf{T}}=\sum_{i\in[N]}(\bm{AY}_{i})^{\circ3}\bm{Y}_{i}^{\mathsf{T}}$,
\item Scale-Invariant: $\mathcal{P}_{\mathrm{O}(n)}(\bm{X})=\mathcal{P}_{\mathrm{O}(n)}(\alpha\bm{X})$,
$\forall\bm{X}\in\mathbb{R}^{n\times n}$, $\forall\alpha>0$.
\end{itemize}
Benefiting from two aforementioned properties, we can know if each
node has the same initial local dictionary and consensus averaging
achieves perfect consensus in each iteration, DMSP will output the
same estimated dictionary as MSP. 

It is worth mentioning that the scale-invariant of projection onto
the orthogonal group $\mathrm{O}(n)$ is a quite benign property not
only because the estimated average gradient matrices obtained by consensus
averaging can be used directly, but also because the auxiliary variable
used in the Push-Sum protocol \cite{kempe2003gossip} no longer needs
to be maintained when the graph is directed, which makes this algorithm
less computational and communication intensive. To be more specific,
because it is costly to construct a doubly stochastic weighted adjacency
matrix $\bm{W}^{(t,s)}$ for a directed graph \cite{gharesifard2012distributed},
consensus averaging in a directed graph is usually through the Push-Sum
protocol, where the weighted adjacency matrix $\bm{W}^{(t,s)}$ only
needs to be column stochastic. In the Push-Sum protocol, the estimated
average variable needs to be divided by an auxiliary variable to get
the correct estimated average variable. However, benefit from the
scale-invariant property, dividing or not dividing by the auxiliary
variable has no effect on the result after projection, so we can dispense
the auxiliary variable.

\begin{algorithm}[tbh]
\caption{DMSP (Simplified Version) \label{alg:alg2}}

1: \textbf{Initialize $\bm{A}_{1}^{(0)}=\bm{A}_{2}^{(0)}=\cdots=\bm{A}_{N}^{(0)}\in\mathrm{O}(n)$}

2: \textbf{for $t=0,1,\ldots,T-1$ do}

3: $\quad\partial\bm{A}_{i}^{(t)}\gets4\left(\bm{A}_{i}^{(t)}\bm{Y}_{i}\right)^{\circ3}\bm{Y}_{i}^{\mathsf{T}},\ \forall i\in[N]$

4: $\quad\left(\partial\bm{\bar{A}}_{i}^{(t)}\right)_{i\in[N]}\gets\textrm{ConsensusAveraging}\left(\left(\partial\bm{A}_{i}^{(t)}\right)_{i\in[N]}\right)$

5: $\quad\bm{U}_{i}\bm{\Sigma}_{i}\bm{V}_{i}^{\mathsf{T}}\gets\mathrm{SVD}\left(\partial\bm{\bar{A}}_{i}^{(t)}\right),\ \forall i\in[N]$

6: $\quad\bm{A}_{i}^{(t+1)}\gets\bm{U}_{i}\bm{V}_{i}^{\mathsf{T}},\ \forall i\in[N]$

7: \textbf{end for}

8: \textbf{Output $\bm{A}_{1}^{(T)},\bm{A}_{2}^{(T)},\ldots,\bm{A}_{N}^{(T)}$}
\end{algorithm}

Since the auxiliary variable can be dispensed in the Push-Sum protocol,
DMSP can use the same algorithmic framework (see Algorithm \ref{alg:alg3})
for both directed and undirected graphs. For undirected graphs, the
in-neighborhood and the out-neighborhood are the same, define the
neighborhood of node $i$ at the $s^{\text{th}}$ inner iteration
in the $t^{\text{th}}$ outer iteration as $\mathcal{N}_{i}^{(t,s)}=\mathcal{N}_{i,in}^{(t,s)}$.
The weighted adjacency matrix $\bm{W}^{(t,s)}$ for undirected graphs
can be constructed according to the Metropolis weights \cite{xiao2004fast,xiao2005a}

\begin{equation}
w_{i,j}^{(t,s)}=\begin{cases}
\frac{1}{\max\left\{ \left|\mathcal{N}_{i}^{(t,s)}\right|,\left|\mathcal{N}_{j}^{(t,s)}\right|\right\} }, & \text{if}\ (i,j)\in\mathcal{E}^{(t,s)},\\
1-\sum_{k\in\mathcal{N}_{i,in}^{(t,s)}\backslash\{i\}}w_{i,k}^{(t,s)}, & \text{if}\ i=j,\\
0, & \text{otherwise},
\end{cases}\label{eq:eq32}
\end{equation}

\noindent while for directed graphs, the weighted adjacency matrix
$\bm{W}^{(t,s)}$ can be constructed according to

\begin{equation}
w_{i,j}^{(t,s)}=\begin{cases}
\frac{1}{\left|\mathcal{N}_{j,out}^{(t,s)}\right|}, & \text{if}\ j\in\mathcal{N}_{i,in}^{(t,s)},\\
0, & \text{otherwise.}
\end{cases}\label{eq:eq33}
\end{equation}

\begin{algorithm}[tbh]
\caption{DMSP (Full Version) \label{alg:alg3}}

$\thinspace$1: \textbf{Initialize $\bm{A}_{1}^{(0)}=\bm{A}_{2}^{(0)}=\cdots=\bm{A}_{N}^{(0)}\in\mathrm{O}(n)$}

$\thinspace$2: \textbf{for $t=0,1,\ldots,T-1$ do}

$\thinspace$3: $\quad\partial\bm{A}_{i}^{(t)}\gets4\left(\bm{A}_{i}^{(t)}\bm{Y}_{i}\right)^{\circ3}\bm{Y}_{i}^{\mathsf{T}},\ \forall i\in[N]$

$\thinspace$4:$\quad\partial\bm{A}_{i}^{(t,0)}\gets\partial\bm{A}_{i}^{(t)},\ \forall i\in[N]$

$\thinspace$5:$\quad$\textbf{for $t_{c}=0,1,\ldots,T_{c}-1$ do}

$\thinspace$6:$\quad\quad\partial\bm{A}_{i}^{\left(t,t_{c}+1\right)}\gets\sum_{j\in\mathcal{N}_{i,in}^{(t)}}w_{i,j}^{\left(t,t_{c}+1\right)}\partial\bm{A}_{j}^{\left(t,t_{c}+1\right)},\ \forall i\in[N]$

$\thinspace$7:$\quad$\textbf{end for}

$\thinspace$8: $\quad\partial\bm{\bar{A}}_{i}^{(t)}\gets\partial\bm{A}_{i}^{\left(t,T_{c}\right)},\ \forall i\in[N]$

$\thinspace$9: $\quad\bm{U}_{i}\bm{\Sigma}_{i}\bm{V}_{i}^{\mathsf{T}}\gets\mathrm{SVD}\left(\partial\bm{\bar{A}}_{i}^{(t)}\right),\ \forall i\in[N]$

10: $\quad\bm{A}_{i}^{(t+1)}\gets\bm{U}_{i}\bm{V}_{i}^{\mathsf{T}},\ \forall i\in[N]$

11: \textbf{end for}

12: \textbf{Output $\bm{A}_{1}^{(T)},\bm{A}_{2}^{(T)},\ldots,\bm{A}_{N}^{(T)}$}
\end{algorithm}

\section{Convergence Analysis}

In practical scenarios, the consensus averaging iteration usually
cannot achieve perfect consensus due to the running time limitation.
After each iteration, the locally estimated dictionary obtained by
each node will have a deviation from the one obtained by MSP because
perfect consensus usually cannot be achieved, and this deviation will
accumulate over the iteration process and affect the DL performance
in subsequent iterations. To analyze whether the dictionaries \textbf{$\bm{A}_{1}^{(T)},\bm{A}_{2}^{(T)},\ldots,\bm{A}_{N}^{(T)}$}
learned by DMSP can approach the dictionary learned by MSP, it is
important to quantify the deviation between the dictionaries learned
by DMSP and the dictionary learned by MSP.

Let $\delta^{(t)}=\max_{i\in[N]}\left\Vert \bm{A}^{(t)}-\bm{A}_{i}^{(t)}\right\Vert _{F}$
denote the deviation of the dictionaries learned by DMSP from the
dictionary learned by MSP after $t$ iterations, and let $\delta_{c}^{(t)}\left(t_{c}^{(t)}\right)=\max_{i\in[N]}\left\Vert \boldsymbol{A}_{i}^{(t)}\left(t_{c}^{(t)}\right)-\bar{\boldsymbol{A}}^{(t)}\right\Vert _{F}$
denote the deviation of dictionaries learned by DMSP from the dictionary
learned by DMSP if the perfect consensus is achieved at the $t^{\text{th}}$
iteration, where $t_{c}^{(t)}$ is the number of consensus averaging
iterations in the $t^{\text{th}}$ iteration of DMSP. We have the
following convergence result of DMSP. 
\begin{thm}[Convergence of DMSP, Informal Version of Theorem \ref{thm:thm6}]
\label{thm:thm1}Suppose that there exists a signed permutation matrix
$\bm{P}$ and a non-negative real number $\epsilon\in\left[0,\frac{\alpha-\theta}{2\alpha n(1-\theta)+3\sqrt{2}(1+\alpha)(1-\theta)+\alpha\theta}\right)$
such that

\begin{align}
\left\Vert \bm{A}^{(t)}-\bm{PD}_{o}^{\mathsf{T}}\right\Vert _{F} & \leq\epsilon,\\
\left\Vert \bm{A}_{i}^{(t)}-\bm{PD}_{o}^{\mathsf{T}}\right\Vert _{F} & \leq\epsilon,\ \forall i\in[N],\\
\left\Vert \bm{A}^{(t)}\bm{D}_{o}\right\Vert _{4}^{4} & \geq n-\frac{1}{2}\left(\epsilon-\delta^{(t)}\right)^{2},
\end{align}

\noindent where constant $\alpha\in(\theta,1)$. If 
\begin{equation}
\delta_{c}^{(k+1)}\in\left[0,\alpha\delta^{(k)}-\frac{2\left\Vert \bm{Z}^{(k)}-\sum_{i\in[N]}\bm{Z}_{i}^{(k)}\right\Vert _{F}}{\sigma_{n}\left(\bm{Z}^{(k)}\right)+\sigma_{n}\left(\sum_{i\in[N]}\bm{Z}_{i}^{(k)}\right)}\right)
\end{equation}

\noindent and $\delta^{(k)}>0$ for every $k\geq t$, where $\bm{Z}^{(k)}=\left(\bm{A}^{(k)}\bm{Y}\right)^{\circ3}\bm{Y}^{\mathsf{T}}$and
$\bm{Z}_{i}^{(k)}=\left(\bm{A}_{i}^{(k)}\bm{Y}_{i}\right)^{\circ3}\bm{Y}_{i}^{\mathsf{T}}$,
then
\begin{equation}
\delta^{(k+1)}<\alpha\delta^{(k)},\ \forall k\geq t
\end{equation}

\noindent holds with high probability, when $p$ is large enough.
\end{thm}
Theorem \ref{thm:thm1} shows that the deviation of the dictionaries
learned by DMSP from the dictionary learned by MSP will converge to
0 at a linear rate with high probability if the estimated dictionaries
are within a neighborhood of a global optimum and the number of consensus
averaging iterations is regulated properly in each iteration. Later
in this section, we will present the formal version of Theorem \ref{thm:thm1}
and prove it rigorously.

Theorem \ref{thm:thm1} takes into account the $t^{\text{th}}$ and
all subsequent iterations. To help our analysis, we start by analyzing
the relationship between $\delta^{(t+1)}$ and $\delta^{(t)}$. As
mentioned before, the deviation between the locally estimated dictionary
obtained by each node and the estimated dictionary obtained by MSP
at the $t+1^{\text{th}}$ iteration can be decomposed into two parts:

\begin{align}
\delta^{(t+1)} & =\max_{i\in[N]}\left\Vert \bm{A}_{i}^{(t+1)}-\bm{A}^{(t+1)}\right\Vert _{F}\nonumber \\
 & \leq\underbrace{\max_{i\in[N]}\left\Vert \bm{A}_{i}^{(t+1)}-\bar{\bm{A}}^{(t+1)}\right\Vert _{F}}_{\delta_{c}^{(t+1)}}+\underbrace{\left\Vert \bar{\bm{A}}^{(t+1)}-\bm{A}^{(t+1)}\right\Vert _{F}}_{\delta_{a}^{(t+1)}},\label{eq:eq30}
\end{align}

\noindent where $\bar{\bm{A}}^{(t+1)}$ is the estimated dictionary
learned by DMSP if the perfect consensus is achieved at the $t+1^{\text{th}}$
iteration. 

(\ref{eq:eq30}) gives an upper bound of $\delta^{(t+1)}$, so we
only need to analyze the relationship between $\delta_{c}^{(t+1)}+\delta_{a}^{(t+1)}$
and $\delta^{(t)}$. In the following, we will first discuss the relationship
between $\delta_{a}^{(t+1)}$ and $\delta^{(t)}$ when the estimated
dictionaries are both within a neighborhood of a global optimum. Since
consensus averaging usually cannot achieve perfect consensus in practical
scenarios, we only consider the case $\delta^{(t)}>0$ without loss
of generality. By the Theorem 1 in \cite{li1995new}, the orthogonal
invariance of the Frobenius norm, and the orthogonal invariance of
singular values, we know that

\begin{equation}
\delta_{a}^{(t+1)}\leq\frac{2\left\Vert \bm{B}^{(t)}-\sum_{i\in[N]}\bm{B}_{i}^{(t)}\right\Vert _{F}}{\sigma_{n}\left(\bm{B}^{(t)}\right)+\sigma_{n}\left(\sum_{i\in[N]}\bm{B}_{i}^{(t)}\right)},\label{eq:eq31}
\end{equation}

\noindent where $\bm{B}^{(t)}=\left(\bm{U}^{(t)}\bm{X}\right)^{\circ3}\bm{X}^{\mathsf{T}}$,
$\bm{U}^{(t)}=\bm{A}^{(t)}\bm{D}_{o}$, $\bm{B}_{i}^{(t)}=\left(\bm{U}_{i}^{(t)}\bm{X}_{i}\right)^{\circ3}\bm{X}_{i}^{\mathsf{T}}$,
$\bm{U}_{i}^{(t)}=\bm{A}_{i}^{(t)}\bm{D}_{o}$, $\forall i\in[N]$.

(\ref{eq:eq31}) implies that we only need to analyze the relationship
between $\frac{2\left\Vert \bm{B}^{(t)}-\sum_{i\in[N]}\bm{B}_{i}^{(t)}\right\Vert _{F}}{\sigma_{n}\left(\bm{B}^{(t)}\right)+\sigma_{n}\left(\sum_{i\in[N]}\bm{B}_{i}^{(t)}\right)}$
and $\delta^{(t)}$. There are two main difficulties in analyzing
this relationship: $\bm{X}$ is a random matrix making we can only
express their relationship in terms of probability, and it is also
tough to obtain a lower bound of $\sigma_{n}(\cdot)$. To solve the
first difficulty, we will first consider the deterministic case that
all random variables are equal to their expectation, and then concentration
bounds can be used. We leverage a gersgorin-type lower bound for the
smallest singular value to solve the second difficulty \cite{johnson1989gersgorin}.
\begin{thm}
\label{thm:thm2} If there exists a signed permutation matrix $\bm{P}\in\mathbb{R}^{n\times n}$
and a non-negative real number $\epsilon\in\left[0,\frac{\alpha-\theta}{2\alpha n(1-\theta)+3\sqrt{2}(1-\alpha)(1+\theta)+\alpha\theta}\right)$,
such that 

\begin{align}
\left\Vert \bm{U}^{(t)}-\bm{P}\right\Vert _{F} & \leq\epsilon,\\
\left\Vert \bm{U}_{i}^{(t)}-\bm{P}\right\Vert _{F} & \leq\epsilon,\ \forall i\in[N],
\end{align}

\noindent where constant $\alpha\in(\theta,1)$, then

\begin{equation}
\frac{2\left\Vert \mathbb{E}\left[\bm{B}^{(t)}\right]-\mathbb{E}\left[\sum_{i\in[N]}\bm{B}_{i}^{(t)}\right]\right\Vert _{F}}{\sigma_{n}\left(\mathbb{E}\left[\bm{B}^{(t)}\right]\right)+\sigma_{n}\left(\mathbb{E}\left[\sum_{i\in[N]}\bm{B}_{i}^{(t)}\right]\right)}<\alpha\delta^{(t)}.\label{eq:eq25}
\end{equation}
\end{thm}
Proof of Theorem \ref{thm:thm2} is given in Appendix \ref{sec:prf2}.
We will next use Theorem \ref{thm:thm2} and concentration bounds
to derive a lower bound of the probability of $\delta_{a}^{(t+1)}<\alpha\delta^{(t)}$
under some assumptions in a stochastic scenario.
\begin{thm}
\label{thm:thm3} On the basis of satisfying all the assumptions in
Theorem \ref{thm:thm2}, if $\frac{p}{(\ln p)^{4}}=\omega\left(\frac{\delta^{(t)}+2}{\delta^{(t)}C}n\right)$,
then

\begin{align}
 & \mathbb{P}\left(\frac{2\left\Vert \bm{B}^{(t)}-\sum_{i\in[N]}\bm{B}_{i}^{(t)}\right\Vert _{F}}{\sigma_{n}\left(\bm{B}^{(t)}\right)+\sigma_{n}\left(\sum_{i\in[N]}\bm{B}_{i}^{(t)}\right)}<\alpha\delta^{(t)}\right)\nonumber \\
\geq & 1-8np\theta\exp\left(-\frac{(\ln p)^{2}}{2}\right)\nonumber \\
- & 8n^{2}\exp\left(\frac{-3p\left(\delta^{(t)}\right)^{2}C^{2}}{c_{2}\left(\delta^{(t)}+2\right)^{2}n^{2}\theta+16n^{\frac{5}{2}}(\ln p)^{4}\delta^{(t)}\left(\delta^{(t)}+2\right)C}\right),\label{eq:eq26}
\end{align}

\noindent where $C=6\alpha\theta(1-\theta)(1-2n\epsilon)-6(1-\alpha)\theta^{2}-6\alpha\theta^{2}\epsilon-18\sqrt{2}(\alpha+1)\theta(1-\theta)\epsilon$
and a constant $c_{2}>6.8\times10^{4}$.
\end{thm}
Proof of Theorem \ref{thm:thm3} is given in Appendix \ref{sec:prf3}.
Now we know the relationship between $\delta_{a}^{(t+1)}$ and $\delta^{(t)}$,
we need to take $\delta_{c}^{(t+1)}$ into account to obtain the relationship
between $\delta_{c}^{(t+1)}+\delta_{a}^{(t+1)}$ and $\delta^{(t)}$.
\begin{thm}
\label{thm:thm4} On the basis of satisfying all the assumptions in
Theorem \ref{thm:thm3}, if 

\noindent 
\begin{equation}
\delta_{c}^{(t+1)}\in\left[0,\alpha\delta^{(t)}-\frac{2\left\Vert \bm{B}^{(t)}-\sum_{i\in[N]}\bm{B}_{i}^{(t)}\right\Vert _{F}}{\sigma_{n}\left(\bm{B}^{(t)}\right)+\sigma_{n}\left(\sum_{i\in[N]}\bm{B}_{i}^{(t)}\right)}\right),
\end{equation}

\noindent then 

\begin{align}
 & \mathbb{P}\left(\delta^{(t+1)}<\alpha\delta^{(t)}\right)\nonumber \\
\geq & 1-8np\theta\exp\left(-\frac{(\ln p)^{2}}{2}\right)\nonumber \\
- & 8n^{2}\exp\left(\frac{-3p\left(\delta^{(t)}\right)^{2}C^{2}}{c_{2}\left(\delta^{(t)}+2\right)^{2}n^{2}\theta+16n^{\frac{5}{2}}(\ln p)^{4}\delta^{(t)}\left(\delta^{(t)}+2\right)C}\right),\label{eq:eq27}
\end{align}

\noindent where $C=6\alpha\theta(1-\theta)(1-2n\epsilon)-6(1-\alpha)\theta^{2}-6\alpha\theta^{2}\epsilon-18\sqrt{2}(\alpha+1)\theta(1-\theta)\epsilon$
and a constant $c_{2}>6.8\times10^{4}$.
\end{thm}
Proof of Theorem \ref{thm:thm4} is given in Appendix \ref{sec:prf4}.
On the basis of satisfying all the assumptions in Theorem \ref{thm:thm4},
if $\frac{p}{(\ln p)^{4}}=\omega\left(\frac{\left(\delta^{(t)}+2\right)^{2}n^{2}\theta}{\left(\delta^{(t)}\right)^{2}C^{2}}+\frac{\left(\delta^{(t)}+2\right)n^{5/2}}{\delta^{(t)}C}\right)$,
we can know that $\delta^{(t+1)}<\alpha\delta^{(t)}$ with high probability.
Based on this conclusion, the relationship between $\delta^{(t+2)}$
and $\delta^{(t+1)}$ will be discussed below.
\begin{thm}
\label{thm:thm5} On the basis of satisfying all the assumptions in
Theorem \ref{thm:thm3}, if

\begin{align}
\left\Vert \bm{U}^{(t)}\right\Vert _{4}^{4} & \geq n-\frac{1}{2}\left(\epsilon-\delta^{(t)}\right)^{2},\label{eq:eq28}\\
\delta^{(t+1)} & <\delta^{(t)},\label{eq:eq29}
\end{align}

\noindent then

\begin{align}
\left\Vert \bm{U}^{(t+1)}-\bm{P}\right\Vert _{F} & \leq\epsilon-\delta^{(t+1)},\\
\left\Vert \bm{U}_{i}^{(t+1)}-\bm{P}\right\Vert _{F} & \leq\epsilon,\ \forall i\in[N],\\
\left\Vert \bm{U}^{(t+1)}\right\Vert _{4}^{4} & \geq n-\frac{1}{2}\left(\epsilon-\delta^{(t+1)}\right)^{2}.
\end{align}
\end{thm}
Proof of Theorem \ref{thm:thm5} is given in Appendix \ref{sec:prf5}.
By Theorem \ref{thm:thm5}, we know that if $\frac{p}{(\ln p)^{4}}=\omega\left(\frac{\left(\delta^{(t+1)}+2\right)^{2}n^{2}\theta}{\left(\delta^{(t+1)}\right)^{2}C^{2}}+\frac{\left(\delta^{(t+1)}+2\right)n^{5/2}}{\delta^{(t+1)}C}\right)$
and $\delta_{c}^{(t+2)}\in\left[0,\alpha\delta^{(t+1)}-\frac{2\left\Vert \bm{B}^{(t+1)}-\sum_{i\in[N]}\bm{B}_{i}^{(t+1)}\right\Vert _{F}}{\sigma_{n}\left(\bm{B}^{(t+1)}\right)+\sigma_{n}\left(\sum_{i\in[N]}\bm{B}_{i}^{(t+1)}\right)}\right)$
also hold, then $\delta^{(t+2)}<\alpha\delta^{(t+1)}$ with high probability
and assumptions in Theorem \ref{thm:thm5} still hold. Therefore,
we can obtain the following theorem.
\begin{thm}[Convergence of DMSP]
\label{thm:thm6} Suppose that there exists a signed permutation
matrix $\bm{P}$ and a non-negative real number $\epsilon\in\left[0,\frac{\alpha-\theta}{2\alpha n(1-\theta)+3\sqrt{2}(1+\alpha)(1-\theta)+\alpha\theta}\right)$
such that

\begin{align}
\left\Vert \bm{A}^{(t)}-\bm{PD}_{o}^{\mathsf{T}}\right\Vert _{F} & \leq\epsilon,\\
\left\Vert \bm{A}_{i}^{(t)}-\bm{PD}_{o}^{\mathsf{T}}\right\Vert _{F} & \leq\epsilon,\ \forall i\in[N],\\
\left\Vert \bm{A}^{(t)}\bm{D}_{o}\right\Vert _{4}^{4} & \geq n-\frac{1}{2}\left(\epsilon-\delta^{(t)}\right)^{2},
\end{align}

\noindent where constant $\alpha\in(\theta,1)$. If 
\begin{equation}
\delta_{c}^{(k+1)}\in\left[0,\alpha\delta^{(k)}-\frac{2\left\Vert \bm{B}^{(k)}-\sum_{i\in[N]}\bm{B}_{i}^{(k)}\right\Vert _{F}}{\sigma_{n}\left(\bm{B}^{(k)}\right)+\sigma_{n}\left(\sum_{i\in[N]}\bm{B}_{i}^{(k)}\right)}\right).
\end{equation}

\noindent and $\delta^{(k)}>0$ for every $k\geq t$, then
\begin{equation}
\delta^{(k+1)}<\alpha\delta^{(k)},\ \forall k\geq t
\end{equation}

\noindent holds with high probability, when $\frac{p}{(\ln p)^{4}}=\omega\left(\frac{\left(\delta^{(k)}+2\right)^{2}n^{2}\theta}{\left(\delta^{(k)}\right)^{2}C^{2}}+\frac{\left(\delta^{(k)}+2\right)n^{5/2}}{\delta^{(k)}C}\right)$.
\end{thm}

\section{Numerical Results}

In this section, the results of comprehensive numerical experiments
will be presented. In Subsection \ref{subsec:5.1}, DMSP will be compared
with MSP to show the effectiveness of DMSP. In Subsection \ref{subsec:5.2},
DMSP will be compared with three existing decentralized DL algorithms
Cloud K-SVD \cite{raja2016cloud}, Linearized D\textsuperscript{4}L
\cite{daneshmand2019decentralized}, and DESTINY \cite{wang2022decentralized}
to further corroborate the efficiency and effectiveness of DMSP. All
experiments are conducted on a server with two Intel Xeon Gold 6354
CPUs. All codes are implemented in \texttt{python}, and \texttt{mpi4py}
\cite{dalcin2005mpi} is used for node-to-node communications. All
experimental results in this section are averaged among 5 trials except
for denoised images.

\subsection{Comparison with MSP \label{subsec:5.1}}

We first consider a time-varying directed network comprising $N=36$
nodes and the edges of this network are randomly generated by an Erdos-Renyi
model with probability parameter $P$. Observation matrix $\bm{Y}\in\mathbb{R}^{n\times p}$
is generated by $\bm{Y}=\bm{D}_{o}\bm{X}$, where $\bm{D}_{o}\in\mathbb{R}^{n\times n}$
is a random orthogonal matrix and $\bm{X}\in\mathbb{R}^{n\times p}$
is a Bernoulli-Gaussian random matrix with sparsity $\theta$. Local
observation matrices $\bm{Y}_{i}$ are obtained by slicing the observation
matrix $\bm{Y}$ approximately evenly. Decentralized nodes collaboratively
recover the dictionary by exchanging the information among neighboring
nodes and the recovery error in each node is measured by $\left|1-\Vert\bm{A}_{i}\bm{D}_{o}\Vert_{4}^{4}/n\right|$because
a perfect recovery gives a 0\% error. The number of outermost iterations
for all experiments in this subsection is 15 and the weighted adjacency
matrices $\bm{W}^{(t,s)}$ are constructed according to (\ref{eq:eq33}).

\begin{figure}[tbh]
\begin{centering}
\includegraphics[scale=0.19]{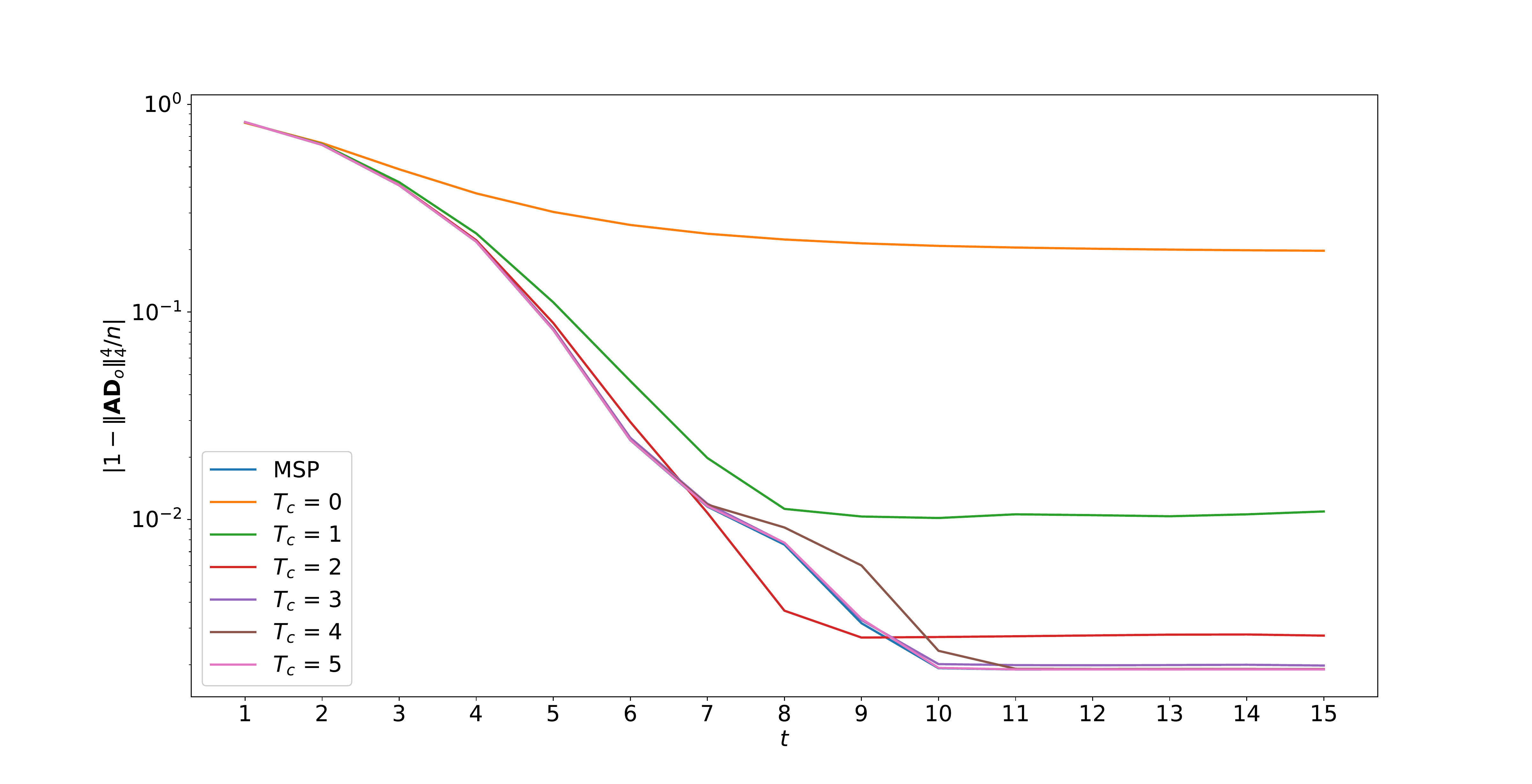}
\par\end{centering}
\caption{\label{fig:fig3} Convergence plots under different choices of $T_{c}$
when $n=25$, $p=10,000$, $\theta=0.1$, and $P=0.2$.}
\end{figure}

Table \ref{tab:tab1} compares the recovery error of MSP and DMSP
under different choices of $P$ and $T_{c}$. Figure \ref{fig:fig3}
shows the convergence performance of MSP and DMSP under different
choices of $T_{c}$. Table \ref{tab:tab1} and Figure \ref{fig:fig3}
show that DMSP converges to MSP within a small number of consensus
averaging iterations per outermost iteration even in a sparse network.
From these experimental results, it can be seen that DMSP can converge
to MSP under rather mild conditions and the effectiveness of DMSP
is verified.

\begin{table*}[tbh]
\caption{\label{tab:tab1}Recovery error under different choices of $P$ and
$T_{c}$.}

\centering{}%
\begin{tabular}{|cccc|c|c|c|c|c|c|c|}
\hline 
$n$ & $p$ & $\theta$ & $P$ & $T_{c}=0$ & $T_{c}=1$ & $T_{c}=2$ & $T_{c}=3$ & $T_{c}=4$ & $T_{c}=5$ & MSP\tabularnewline
\hline 
25 & 10,000 & 0.1 & 0.2 & 18.50\% & 1.03\% & 0.27\% & 0.20\% & 0.19\% & 0.19\% & 0.19\%\tabularnewline
25 & 10,000 & 0.1 & 0.5 & 18.50\% & 0.37\% & 0.20\% & 0.19\% & 0.19\% & 0.19\% & 0.19\%\tabularnewline
25 & 10,000 & 0.1 & 0.8 & 18.50\% & 0.24\% & 0.19\% & 0.19\% & 0.19\% & 0.19\% & 0.19\%\tabularnewline
\hline 
\end{tabular}
\end{table*}

\subsection{Comparison with Existing Decentralized DL Algorithms \label{subsec:5.2}}

In this subsection, DMSP will be compared with three existing decentralized
DL algorithms Cloud K-SVD \cite{raja2016cloud}, Linearized D\textsuperscript{4}L
\cite{daneshmand2019decentralized}, and DESTINY \cite{wang2022decentralized},
using both synthetic and real data. Cloud K-SVD, Linearized D\textsuperscript{4}L,
and DESTINY are chosen as baselines because they are representatives
of $\ell^{0}$-based, $\ell^{1}$-based, and $\ell^{4}$-based decentralized
DL algorithms, respectively. To meet the requirements of all the baselines
being compared, we consider a time-invariant undirected network comprising
$N=36$ nodes and the edges of this network are randomly generated
by an Erdos-Renyi model with probability parameter $P=0.5$. To ensure
that the communication cost of each outermost iteration of these four
algorithms is approximately the same, let $T_{c}=2$ in DMSP and $T_{p}=2$,
$T_{c}=1$ in Cloud K-SVD. The weighted adjacency matrices $\bm{W}^{(t,s)}$
are constructed according to (\ref{eq:eq32}).

\subsubsection{Experiments Using Bernoulli-Gaussian Data}

The observation matrix $\bm{Y}\in\mathbb{R}^{n\times p}$ is generated
by $\bm{Y}=\bm{D}_{o}\bm{X}$, where $\bm{D}_{o}\in\mathbb{R}^{n\times n}$
is a random orthogonal matrix and $\bm{X}\in\mathbb{R}^{n\times p}$
is a Bernoulli-Gaussian random matrix with sparsity $\theta$. Local
observation matrices $\bm{Y}_{i}\in\mathbb{R}^{n\times p_{i}}$ are
obtained by slicing the observation matrix $\bm{Y}$ approximately
evenly. The metric for recovery error is the same as the metric in
Subsection \ref{subsec:5.1}.

\begin{figure}[tbh]
\centering{}\includegraphics[scale=0.19]{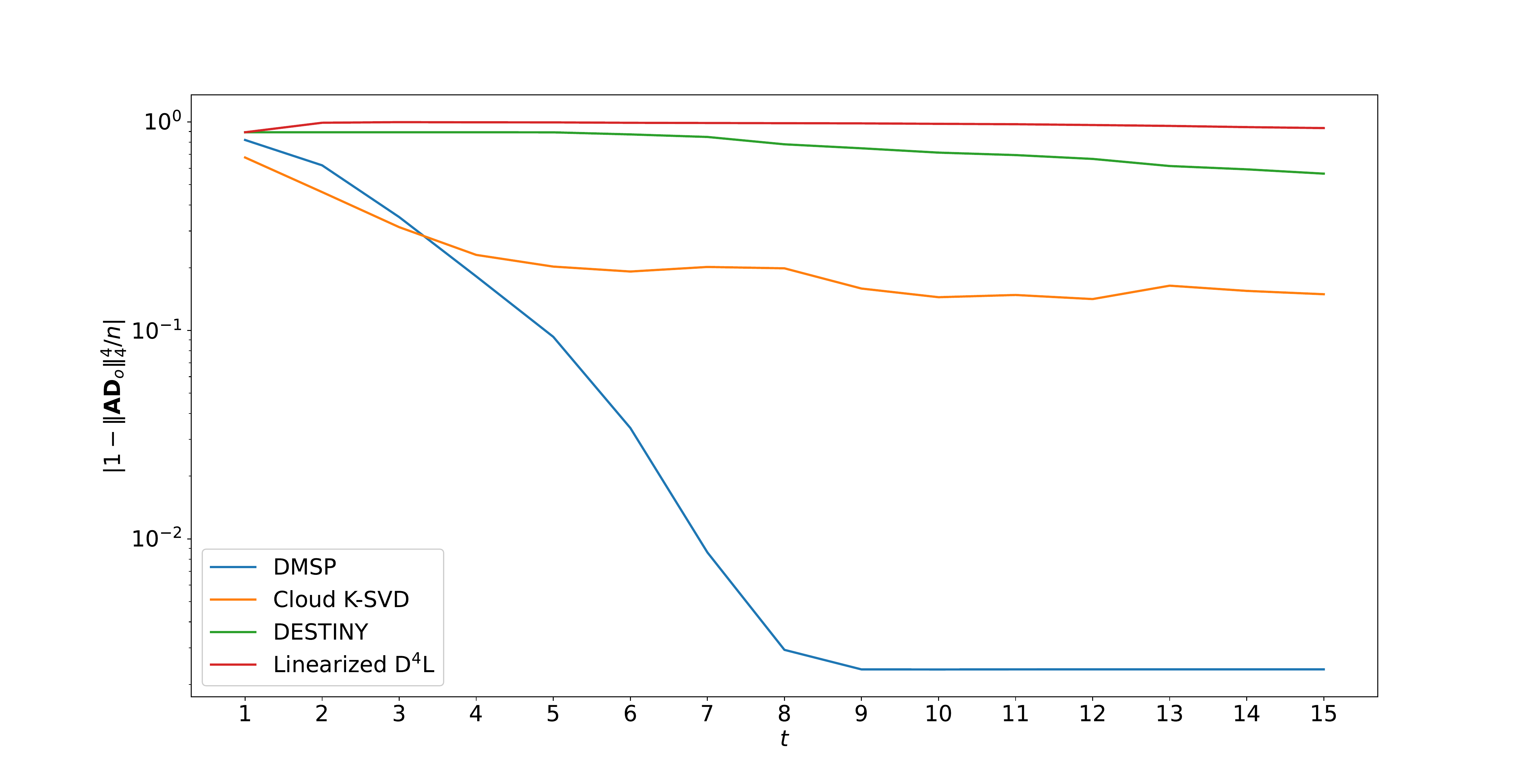}\caption{\label{fig:fig4} Convergence plots when $n=25$, $p=10,000$, and
$\theta=0.1$.}
\end{figure}

Table \ref{tab:tab2} compares DMSP with three baseline algorithms
in terms of the running time and the recovery errors under different
choices of $n$, $p$, and $\theta$. The experimental results in
Table \ref{tab:tab2} show that DMSP has a lower per-iteration computational
complexity and faster convergence rate than existing decentralized
DL algorithms, which also results in a lower communication cost required
to achieve a desired convergence result. Moreover, the low per-iteration
computational complexity and fast convergence rate of DMSP indicate
that it is more suitable for large-scale scenarios than existing decentralized
DL algorithms. Figure \ref{fig:fig4} presents the convergence performance
of DMSP and three existing decentralized DL algorithms mentioned above
when $n=25$, $p=10,000$, and $\theta=0.1$. Figure \ref{fig:fig4}
shows that DMSP converges faster than existing decentralized DL algorithms.

\begin{table*}[tbh]
\begin{centering}
\caption{\label{tab:tab2}Recovery performance under different choices of $n$,
$p$, and $\theta$.}
\par\end{centering}
\centering{}%
\begin{tabular}{|cccc|cc|cc|cc|cc|}
\hline 
 &  &  &  & \multicolumn{2}{c|}{DMSP (Ours)} & \multicolumn{2}{c|}{Cloud K-SVD} & \multicolumn{2}{c|}{DESTINY} & \multicolumn{2}{c|}{Linearized D\textsuperscript{4}L}\tabularnewline
\hline 
$n$ & $p$ & $\theta$ & iter. & Time & Error & Time & Error & Time & Error & Time & Error\tabularnewline
\hline 
25 & 10,000 & 0.1 & 15 & 0.06$\thinspace$s & 0.26\% & 1.18$\thinspace$s & 13.71\% & 0.06$\thinspace$s & 57.89\% & 0.07$\thinspace$s & 94.61\%\tabularnewline
25 & 10,000 & 0.3 & 15 & 0.05$\thinspace$s & 0.38\% & 1.30$\thinspace$s & 13.45\% & 0.05$\thinspace$s & 60.20\% & 0.07$\thinspace$s & 99.73\%\tabularnewline
50 & 20,000 & 0.1 & 20 & 2.42$\thinspace$s & 0.27\% & 107.48$\thinspace$s & 15.84\% & 3.23$\thinspace$s & 63.49\% & 4.78$\thinspace$s & 88.30\%\tabularnewline
50 & 20,000 & 0.3 & 20 & 2.98$\thinspace$s & 0.39\% & 79.38$\thinspace$s & 15.24\% & 3.73$\thinspace$s & 72.23\% & 3.69$\thinspace$s & 97.24\%\tabularnewline
100 & 40,000 & 0.1 & 25 & 8.69$\thinspace$s & 0.25\% & 485.30$\thinspace$s & 18.49\% & 10.33$\thinspace$s & 72.86\% & 10.21$\thinspace$s & 92.36\%\tabularnewline
100 & 40,000 & 0.3 & 25 & 8.54$\thinspace$s & 0.40\% & 499.42$\thinspace$s & 16.28\% & 9.58$\thinspace$s & 81.44\% & 12.25$\thinspace$s & 99.60\%\tabularnewline
\hline 
\end{tabular}
\end{table*}

\subsubsection{Experiments on Image Denoising}

The theoretical analysis of DMSP heavily relies on the Bernoulli-Gaussian
model. Experiments in this part aim to illustrate the performance
of DMSP beyond the Bernoulli-Gaussian model by comparing the effectiveness
and efficiency of DMSP with existing decentralized DL algorithms on
image denoising.

The original image chosen for experiments on image denoising are two
$512\times512$ grayscale images \texttt{Boat} and\texttt{ Barbara}.
The corrupted images are obtained by adding Gaussian white noise to
the original images with different noise levels. For each experiment,
the observation matrix $\bm{Y}$ is generated by extracting all $8\times8$
patches from a corrupted image. Local observation matrices $\bm{Y}_{i}$
are obtained by slicing the observation matrix $\bm{Y}$ approximately
evenly. Peak signal-to-noise ratio (PSNR) is used to measure the effectiveness
of denoising and the number of outermost iteration for all experiments
in this part is 30. 

\begin{figure}[tbh]
\begin{minipage}[t]{0.3\columnwidth}%
\begin{center}
\includegraphics[scale=0.14]{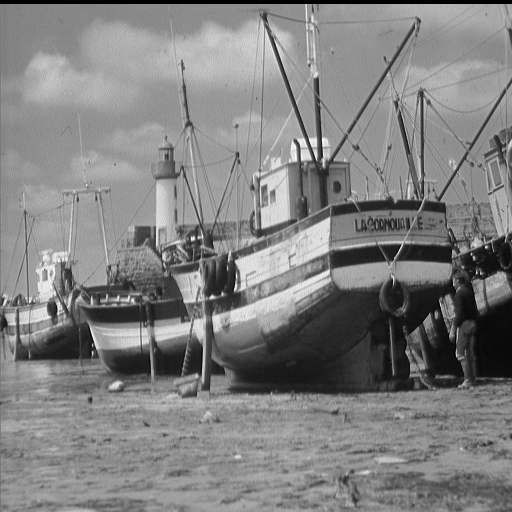}
\par\end{center}
\begin{center}
(a)
\par\end{center}%
\end{minipage}\hfill{}%
\begin{minipage}[t]{0.3\columnwidth}%
\begin{center}
\includegraphics[scale=0.14]{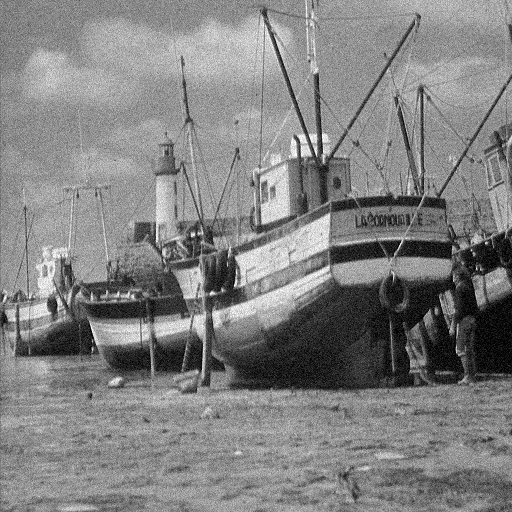}
\par\end{center}
\begin{center}
(b)
\par\end{center}%
\end{minipage}\hfill{}%
\begin{minipage}[t]{0.3\columnwidth}%
\begin{center}
\includegraphics[scale=0.14]{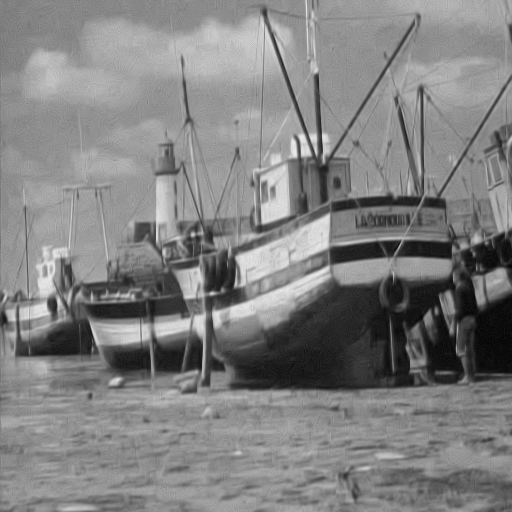}
\par\end{center}
\begin{center}
(c)
\par\end{center}%
\end{minipage}\medskip{}

\begin{minipage}[t]{0.3\columnwidth}%
\begin{center}
\includegraphics[scale=0.14]{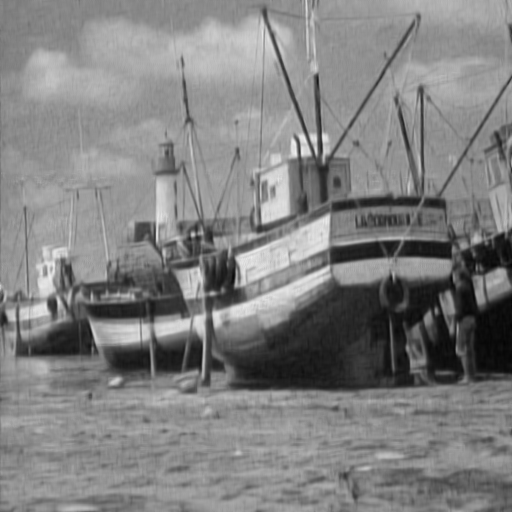}
\par\end{center}
\begin{center}
(d)
\par\end{center}%
\end{minipage}\hfill{}%
\begin{minipage}[t]{0.3\columnwidth}%
\begin{center}
\includegraphics[scale=0.14]{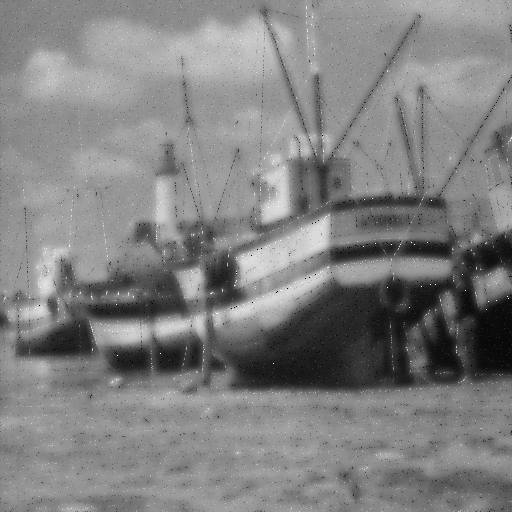}
\par\end{center}
\begin{center}
(e)
\par\end{center}%
\end{minipage}\hfill{}%
\begin{minipage}[t]{0.3\columnwidth}%
\begin{center}
\includegraphics[scale=0.14]{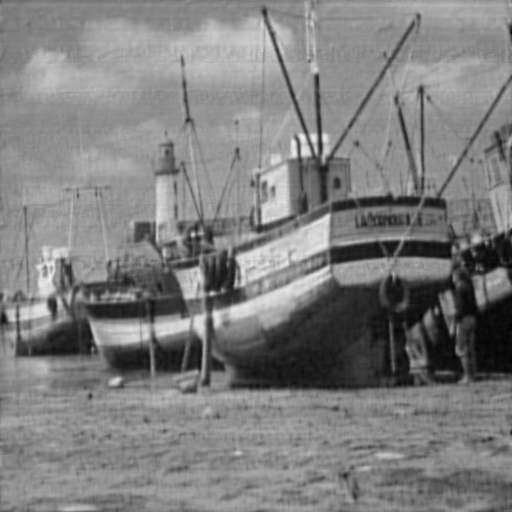}
\par\end{center}
\begin{center}
(f)
\par\end{center}%
\end{minipage}

\caption{\label{fig:fig1}Original image, corrupted image, and denoised images
after 30 outermost iterations of \texttt{Boat}. (a): original image;
(b): corrupted image (PSNR$\thinspace$=$\thinspace$26.04$\thinspace$dB);
(c): denoised image by DMSP (PSNR$\thinspace$=$\thinspace$30.03$\thinspace$dB);
(d): denoised image by Cloud K-SVD (PSNR$\thinspace$=$\thinspace$28.50$\thinspace$dB);
(e): denoised image by DESTINY (PSNR$\thinspace$=$\thinspace$24.28$\thinspace$dB);
(f): denoised image by Linearized D\protect\textsuperscript{4}L (PSNR$\thinspace$=$\thinspace$27.27$\thinspace$dB).}
\end{figure}

\begin{figure}[tbh]
\begin{minipage}[t]{0.3\columnwidth}%
\begin{center}
\includegraphics[scale=0.14]{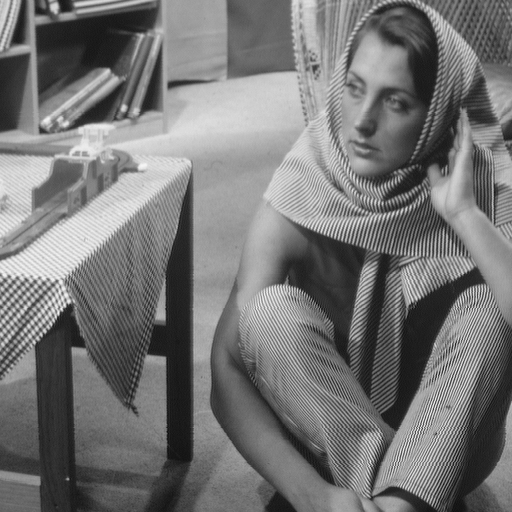}
\par\end{center}
\begin{center}
(a)
\par\end{center}%
\end{minipage}\hfill{}%
\begin{minipage}[t]{0.3\columnwidth}%
\begin{center}
\includegraphics[scale=0.14]{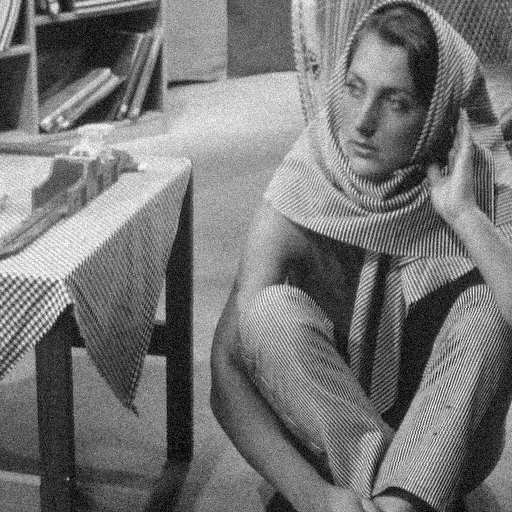}
\par\end{center}
\begin{center}
(b)
\par\end{center}%
\end{minipage}\hfill{}%
\begin{minipage}[t]{0.3\columnwidth}%
\begin{center}
\includegraphics[scale=0.14]{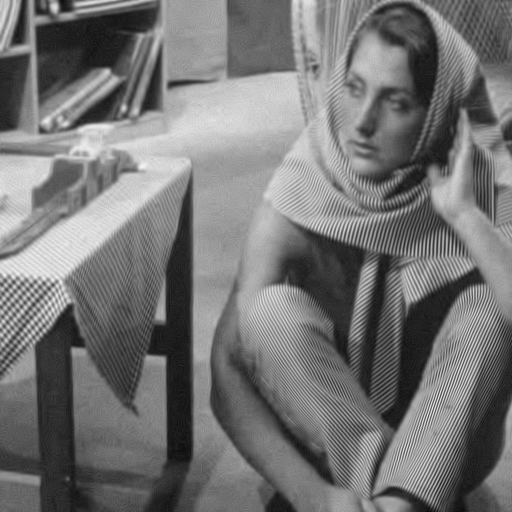}
\par\end{center}
\begin{center}
(c)
\par\end{center}%
\end{minipage}\medskip{}

\begin{minipage}[t]{0.3\columnwidth}%
\begin{center}
\includegraphics[scale=0.14]{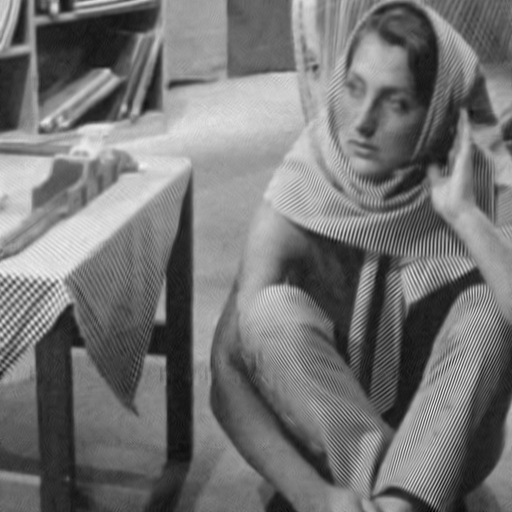}
\par\end{center}
\begin{center}
(d)
\par\end{center}%
\end{minipage}\hfill{}%
\begin{minipage}[t]{0.3\columnwidth}%
\begin{center}
\includegraphics[scale=0.14]{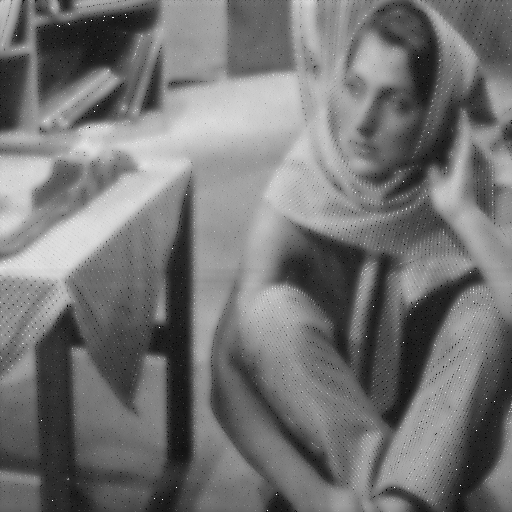}
\par\end{center}
\begin{center}
(e)
\par\end{center}%
\end{minipage}\hfill{}%
\begin{minipage}[t]{0.3\columnwidth}%
\begin{center}
\includegraphics[scale=0.14]{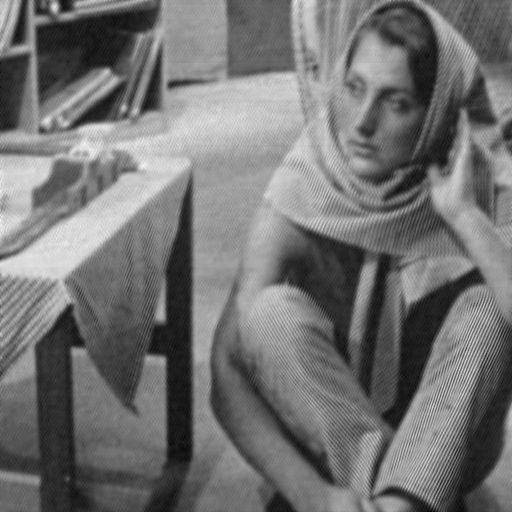}
\par\end{center}
\begin{center}
(f)
\par\end{center}%
\end{minipage}

\caption{\label{fig:fig2}Original image, corrupted image, and denoised images
after 30 outermost iterations of \texttt{Barbara}. (a): original image;
(b): corrupted image (PSNR$\thinspace$=$\thinspace$26.03$\thinspace$dB);
(c): denoised image by DMSP (PSNR$\thinspace$=$\thinspace$30.15$\thinspace$dB);
(d): denoised image by Cloud K-SVD (PSNR$\thinspace$=$\thinspace$28.37$\thinspace$dB);
(e): denoised image by DESTINY (PSNR$\thinspace$=$\thinspace$23.22$\thinspace$dB);
(f): denoised image by Linearized D\protect\textsuperscript{4}L (PSNR$\thinspace$=$\thinspace$25.56$\thinspace$dB).}
\end{figure}

Table \ref{tab:tab3} compares DMSP with three baseline algorithms
in terms of the running time and the effectiveness of denoising under
different choices of noise level. Figure \ref{fig:fig1} and Figure
\ref{fig:fig2} presents denoised images generated by DMSP and three
baseline algorithms when the noise level is 0.0025. The experimental
results in Table \ref{tab:tab3}, Figure \ref{fig:fig1}, and Figure
\ref{fig:fig2} show that DMSP outperforms existing decentralized
DL algorithms on image denoising in terms of the effectiveness of
denoising, per-iteration computational complexity, convergence rate,
and communication cost.

\begin{table*}[tbh]
\caption{\label{tab:tab3}Denoising performance under different choices of
image and noise level.}

\centering{}%
\begin{tabular}{|c|cc|cc|cc|cc|cc|}
\hline 
 & \multicolumn{2}{c|}{Corrupted Image} & \multicolumn{2}{c|}{DMSP (Ours)} & \multicolumn{2}{c|}{Cloud K-SVD} & \multicolumn{2}{c|}{DESTINY} & \multicolumn{2}{c|}{Linearized D\textsuperscript{4}L}\tabularnewline
\hline 
 & Variance & PSNR & Time & PSNR & Time & PSNR & Time & PSNR & Time & PSNR\tabularnewline
\hline 
\multirow{3}{*}{\texttt{Boat}} & 0.0025 & 26.04$\thinspace$dB & 21.47$\thinspace$s & 30.01$\thinspace$dB & 2954.23$\thinspace$s & 28.57$\thinspace$dB & 36.64$\thinspace$s & 24.33$\thinspace$dB & 44.95$\thinspace$s & 27.36$\thinspace$dB\tabularnewline
 & 0.005 & 23.07$\thinspace$dB & 20.31$\thinspace$s & 28.99$\thinspace$dB & 3049.90$\thinspace$s & 28.20$\thinspace$dB & 36.70$\thinspace$s & 23.75$\thinspace$dB  & 46.85$\thinspace$s & 26.79$\thinspace$dB\tabularnewline
 & 0.01 & 20.11$\thinspace$dB & 19.75$\thinspace$s & 27.39$\thinspace$dB & 2833.23$\thinspace$s & 27.35$\thinspace$dB & 37.81$\thinspace$s & 22.94$\thinspace$dB & 46.72$\thinspace$s & 26.57$\thinspace$dB\tabularnewline
\hline 
\multirow{3}{*}{\texttt{Barbara}} & 0.0025 & 26.03$\thinspace$dB & 20.31$\thinspace$s & 30.16$\thinspace$dB & 3097.09$\thinspace$s & 28.27$\thinspace$dB & 36.70$\thinspace$s & 23.07$\thinspace$dB & 46.00$\thinspace$s & 25.82$\thinspace$dB\tabularnewline
 & 0.005 & 23.05$\thinspace$dB & 20.52$\thinspace$s & 29.25$\thinspace$dB & 3058.77$\thinspace$s & 27.68$\thinspace$dB & 38.16$\thinspace$s & 20.63$\thinspace$dB & 45.69$\thinspace$s & 25.40$\thinspace$dB\tabularnewline
 & 0.01 & 20.14$\thinspace$dB & 19.70$\thinspace$s & 27.77$\thinspace$dB & 2917.84$\thinspace$s & 27.12$\thinspace$dB & 36.85$\thinspace$s & 19.82$\thinspace$dB & 45.59$\thinspace$s & 24.73$\thinspace$dB\tabularnewline
\hline 
\end{tabular}
\end{table*}

\section{Conclusion}

In this paper, we proposed a novel decentralized DL algorithm DMSP,
which allows decentralized nodes collaboratively learn the dictionary
from the local data. Convergence analysis showed that DMSP can converge
to MSP at a linear rate with high probability under certain conditions,
which indicates that DMSP can effectively leverage the advantages
of $\ell^{4}$-norm maximization to achieve excellent performance
in terms of per-iteration computational complexity, convergence rate,
etc. Extensive experiments further corroborated the effectiveness
and efficiency of DMSP.

\appendices{}

\section{Lemmas for the Convergence Analysis}
\begin{lem}
\label{lem:lem1} If $\bm{A}\in\mathbb{R}^{n\times n}$, then

\begin{equation}
\sigma_{n}(\bm{A})\geq\min_{i\in[n]}\left\{ \left|a_{i,i}\right|-\frac{1}{2}\left(\sum_{j\neq i}\left|a_{i,j}\right|+\sum_{j\neq i}\left|a_{j,i}\right|\right)\right\} .\label{eq:eq4}
\end{equation}
\end{lem}
\begin{IEEEproof}
\noindent Please refer to Theorem 3 in \cite{johnson1989gersgorin}.
\end{IEEEproof}
\begin{lem}
\label{lem:lem2} If $\bm{U}\in\mathrm{O}(n),\bm{X}\in\mathbb{R}^{n\times p},x_{i,j}\sim_{i.i.d.}\mathrm{BG}(\theta)$,
then

\begin{multline}
\left\{ \mathbb{E}\left[(\bm{UX})^{\circ3}(\bm{UX})^{\mathsf{T}}\right]\right\} _{i,j}\\
=\begin{cases}
3p\theta(1-\theta)\sum_{k\in[n]}u_{i,k}^{3}u_{j,k}, & \text{if}\ i\neq j,\\
3p\theta(1-\theta)\sum_{k\in[n]}u_{i,k}^{4}+3p\theta^{2}, & \text{if}\ i=j.
\end{cases}\label{eq:eq5}
\end{multline}
\end{lem}
\begin{IEEEproof}
\noindent By (98) in \cite{zhai2020complete}, we know that

\begin{equation}
\mathbb{E}\left[(\bm{UX})^{\circ3}\bm{X}^{\mathsf{T}}\right]=3p\theta(1-\theta)\bm{U}^{\circ3}+3p\theta^{2}\bm{U}.
\end{equation}

\noindent Since $\bm{U}$ is deterministic and orthogonal, 

\begin{equation}
\mathbb{E}\left[(\bm{UX})^{\circ3}(\bm{UX})^{\mathsf{T}}\right]=3p\theta(1-\theta)\bm{U}^{\circ3}\bm{U}^{\mathsf{T}}+3p\theta^{2}\bm{I}.
\end{equation}

\noindent Therefore,

\begin{multline}
\left\{ \mathbb{E}\left[(\bm{UX})^{\circ3}(\bm{UX})^{\mathsf{T}}\right]\right\} _{i,j}\\
=\begin{cases}
3p\theta(1-\theta)\sum_{k\in[n]}u_{i,k}^{3}u_{j,k}, & \text{if}\ i\neq j,\\
3p\theta(1-\theta)\sum_{k\in[n]}u_{i,k}^{4}+3p\theta^{2}, & \text{if}\ i=j.
\end{cases}
\end{multline}
\end{IEEEproof}
\begin{lem}
\label{lem:lem4} Let $\bm{P}\in\mathbb{R}^{n\times n}$ be a signed
permutation matrix and $\bm{U}\in\mathrm{O}(n)$. If $\Vert\bm{U}-\bm{P}\Vert_{F}=\epsilon$
and $p_{i,j}\neq0$, then

\begin{equation}
\left|u_{i,j}\right|\geq1-\frac{1}{2}\epsilon^{2}.
\end{equation}
\end{lem}
\begin{IEEEproof}
\noindent We can assume that $\bm{P}$ is a diagonal matrix without
loss of generality. 

\noindent If $p_{i,i}=1$, let $\bm{u}_{i,\ast}=[\delta_{1},\ldots,\delta_{i-1},1-\delta_{i},\delta_{i+1},\ldots,\delta_{n}]$
and $\epsilon_{i}^{2}=\sum_{i\in[n]}\delta_{i}^{2}$. Since $\bm{U}$
is orthogonal, we know that 

\begin{equation}
\delta_{1}^{2}+\cdots+\delta_{i-1}^{2}+(1-\delta_{i})^{2}+\delta_{i+1}^{2}+\cdots+\delta_{n}^{2}=1.\label{eq:eq7}
\end{equation}

\noindent Substitute $\sum_{j\neq i}\delta_{j}^{2}$ by $\epsilon_{i}^{2}-\delta_{i}^{2}$
in (\ref{eq:eq7}), we obtain that

\begin{equation}
(1-\delta_{i})^{2}+\epsilon_{i}^{2}-\delta_{i}^{2}=1.\label{eq:eq8}
\end{equation}

\noindent Expand $(1-\delta_{i})^{2}$ in (\ref{eq:eq8}), we have

\begin{equation}
\delta_{i}=\frac{1}{2}\epsilon_{i}^{2}\leq\frac{1}{2}\epsilon,
\end{equation}

\noindent which indicates that

\begin{equation}
\left|u_{i,i}\right|\geq1-\frac{1}{2}\epsilon^{2}.
\end{equation}

\noindent If $p_{i,i}=-1$, we can get the same result by the same
approach.

\noindent Therefore, we can conclude that if $\Vert\bm{U}-\bm{P}\Vert_{F}=\epsilon$
and $p_{i,j}\neq0$, then

\begin{equation}
\left|u_{i,j}\right|\geq1-\frac{1}{2}\epsilon^{2}.
\end{equation}
\end{IEEEproof}
\begin{lem}
\label{lem:lem5} Let $\bm{P}\in\mathbb{R}^{n\times n}$ be a signed
permutation matrix and $\bm{U}\in\mathrm{O}(n)$. If $\Vert\bm{U}-\bm{P}\Vert_{F}=\epsilon\leq1$
and $i\neq j$, then

\begin{equation}
\left|\bm{u}_{i,\ast}^{\circ3}\bm{u}_{j,\ast}^{\mathsf{T}}\right|\leq2\epsilon.
\end{equation}
\end{lem}
\begin{IEEEproof}
\noindent Let $\theta_{\bm{u}_{i},\bm{u}_{i}^{\circ3}}$, $\theta_{\bm{u}_{i},\bm{u}_{j}}$,
$\theta_{\bm{u}_{i}^{\circ3},\bm{u}_{j}}\in[0,\pi]$ be the angles
between vector $\bm{u}_{i}$ and vector $\bm{u}_{i}^{\circ3}$, vector
$\bm{u}_{i}$ and vector $\bm{u}_{j}$, and vector $\bm{u}_{i}^{\circ3}$
and vector $\bm{u}_{j}$, respectively. The orthogonality of $\bm{U}$
implies that

\begin{equation}
\left|\bm{u}_{i,\ast}^{\circ3}\bm{u}_{j,\ast}^{\mathsf{T}}\right|\leq\left\Vert \bm{u}_{i,\ast}^{\circ3}\right\Vert _{2}\left\Vert \bm{u}_{j,\ast}\right\Vert _{2}\left|\cos\theta_{\bm{u}_{i}^{\circ3},\bm{u}_{j}}\right|\leq\left|\cos\theta_{\bm{u}_{i}^{\circ3},\bm{u}_{j}}\right|.\label{eq:eq9}
\end{equation}

\noindent If $\theta_{\bm{u}_{i}^{\circ3},\bm{u}_{j}}\in\left[0,\frac{\pi}{2}\right]$,
we know that

\begin{equation}
\theta_{\bm{u}_{i}^{\circ3},\bm{u}_{j}}\geq\theta_{\bm{u}_{i},\bm{u}_{j}}-\theta_{\bm{u}_{i},\bm{u}_{i}^{\circ3}}=\frac{\pi}{2}-\theta_{\bm{u}_{i},\bm{u}_{i}^{\circ3}},\label{eq:eq11}
\end{equation}

\noindent which implies that

\begin{equation}
\left|\cos\theta_{\bm{u}_{i}^{\circ3},\bm{u}_{j}}\right|\leq\left|\cos\left(\frac{\pi}{2}-\theta_{\bm{u}_{i},\bm{u}_{i}^{\circ3}}\right)\right|=\sin\theta_{\bm{u}_{i},\bm{u}_{i}^{\circ3}}.\label{eq:eq12}
\end{equation}

\noindent If $\theta_{\bm{u}_{i}^{\circ3},\bm{u}_{j}}\in\left(\frac{\pi}{2},\pi\right]$,
we know that

\begin{equation}
\theta_{\bm{u}_{i}^{\circ3},\bm{u}_{j}}\leq\theta_{\bm{u}_{i},\bm{u}_{j}}+\theta_{\bm{u}_{i},\bm{u}_{i}^{\circ3}}=\frac{\pi}{2}+\theta_{\bm{u}_{i},\bm{u}_{i}^{\circ3}},\label{eq:eq13}
\end{equation}

\noindent which implies that

\begin{equation}
\left|\cos\theta_{\bm{u}_{i}^{\circ3},\bm{u}_{j}}\right|\leq\left|\cos\left(\frac{\pi}{2}+\theta_{\bm{u}_{i},\bm{u}_{i}^{\circ3}}\right)\right|=\sin\theta_{\bm{u}_{i},\bm{u}_{i}^{\circ3}}.\label{eq:eq14}
\end{equation}

\noindent By (\ref{eq:eq9}), (\ref{eq:eq12}), and (\ref{eq:eq14}),
we know that 

\begin{equation}
\left|\bm{u}_{i,\ast}^{\circ3}\bm{u}_{j,\ast}^{\mathsf{T}}\right|\leq\left|\cos\theta_{\bm{u}_{i}^{\circ3},\bm{u}_{j}}\right|\leq\sin\theta_{\bm{u}_{i},\bm{u}_{i}^{\circ3}}.
\end{equation}

\noindent We can further obtain that

\begin{align}
\left|\bm{u}_{i,\ast}^{\circ3}\bm{u}_{j,\ast}^{\mathsf{T}}\right| & \leq\sin\theta_{\bm{u}_{i},\bm{u}_{i}^{\circ3}}\nonumber \\
 & =\sin\left(\arccos\frac{\sum_{k\in[n]}u_{i,k}^{4}}{\left\Vert \bm{u}_{i,\ast}^{\circ3}\right\Vert _{2}\left\Vert \bm{u}_{i,\ast}\right\Vert _{2}}\right)\nonumber \\
 & =\sqrt{1-\left(\frac{\sum_{k\in[n]}u_{i,k}^{4}}{\left\Vert \bm{u}_{i,\ast}^{\circ3}\right\Vert _{2}\left\Vert \bm{u}_{i,\ast}\right\Vert _{2}}\right)^{2}}\nonumber \\
 & \leq\sqrt{1-\left(\sum_{k\in[n]}u_{i,k}^{4}\right)^{2}}\nonumber \\
 & \leq\sqrt{1-\left(1-\frac{1}{2}\epsilon^{2}\right)^{8}}\nonumber \\
 & \leq2\epsilon,
\end{align}
where the third inequality is obtained from Lemma \ref{lem:lem4}.
\end{IEEEproof}
\begin{lem}
\label{lem:lem6} Let $\bm{U}\in\mathrm{O}(n),\bm{X}\in\mathbb{R}^{n\times p},x_{i,j}\sim_{i.i.d.}\mathrm{BG}(\theta)$.
If there exists a signed permutation matrix $\bm{P}\in\mathbb{R}^{n\times n}$,
such that $\Vert\bm{U}-\bm{P}\Vert_{F}=\epsilon\leq1$, then

\begin{equation}
\sigma_{n}\left(\mathbb{E}\left[(\bm{UX})^{\circ3}\bm{X}^{\mathsf{T}}\right]\right)\geq3p\theta(1-\theta)(1-2n\epsilon)+3p\theta^{2}.
\end{equation}
\end{lem}
\begin{IEEEproof}
\noindent By Lemma \ref{lem:lem2} and Lemma \ref{lem:lem4}, we know
that

\begin{align}
\left|\left\{ \mathbb{E}\left[(\bm{UX})^{\circ3}(\bm{UX})^{\mathsf{T}}\right]\right\} _{i,i}\right| & =3p\theta(1-\theta)\sum_{k\in[n]}u_{i,k}^{4}+3p\theta^{2}\nonumber \\
 & \geq3p\theta(1-\theta)\left(1-\frac{1}{2}\epsilon^{2}\right)^{4}+3p\theta^{2}\nonumber \\
 & \geq3p\theta(1-\theta)\left(1-\frac{1}{2}\epsilon\right)^{4}+3p\theta^{2}\nonumber \\
 & \geq3p\theta(1-\theta)\left(1-2\epsilon\right)+3p\theta^{2}.\label{eq:eq15}
\end{align}

\noindent By Lemma \ref{lem:lem2} and Lemma \ref{lem:lem5}, we know
that if $i\neq j$, then

\begin{align}
\left|\left\{ \mathbb{E}\left[(\bm{UX})^{\circ3}(\bm{UX})^{\mathsf{T}}\right]\right\} _{i,j}\right| & =\left|3p\theta(1-\theta)\sum_{k\in[n]}u_{i,k}^{3}u_{j,k}\right|\nonumber \\
 & =3p\theta(1-\theta)\left|\bm{u}_{i,\ast}^{\circ3}\bm{u}_{j,\ast}^{\mathsf{T}}\right|\nonumber \\
 & \leq3p\theta(1-\theta)(2\epsilon).\label{eq:eq16}
\end{align}
Hence, (\ref{eq:eq15}), (\ref{eq:eq16}), and Lemma \ref{lem:lem1}
imply that

\begin{equation}
\sigma_{n}\left(\mathbb{E}\left[(\bm{UX})^{\circ3}(\bm{UX})^{\mathsf{T}}\right]\right)\geq3p\theta(1-\theta)(1-2n\epsilon)+3p\theta^{2}.
\end{equation}

\noindent Finally, by the orthogonal invariance of singular values,
we have
\begin{equation}
\sigma_{n}\left(\mathbb{E}\left[(\bm{UX})^{\circ3}\bm{X}^{\mathsf{T}}\right]\right)\geq3p\theta(1-\theta)(1-2n\epsilon)+3p\theta^{2}.
\end{equation}
\end{IEEEproof}
\begin{lem}
\label{lem:lem7} Let $\bm{U}$, $\bm{V}\in\mathrm{O}(n)$. If there
exists a signed permutation matrix $\bm{P}$, such that $\Vert\bm{U}-\bm{P}\Vert_{F}\leq\epsilon\leq1$
and $\Vert\bm{V}-\bm{P}\Vert_{F}\leq\epsilon\leq1$, then

\begin{equation}
\left\Vert \bm{U}^{\circ3}-\bm{V}^{\circ3}\right\Vert _{F}\leq3\sqrt{2}\epsilon\Vert\bm{U}-\bm{V}\Vert_{F}.\label{eq:eq17}
\end{equation}
\end{lem}
\begin{IEEEproof}
\noindent Let $\delta=\Vert\bm{U}-\bm{V}\Vert_{F}$. If $\delta=0$,
(\ref{eq:eq17}) is trivial, so we only need to consider the case
where $\delta>0$. 

\noindent We can assume that $\bm{P}$ is a diagonal matrix without
loss of generality. Let $\bm{U}-\bm{V}=\bm{D}+\bm{N}$ where $\bm{D}$
is the diagonal part of $\bm{U}-\bm{V}$ and $\bm{N}$ is the off-diagonal
part of $\bm{U}-\bm{V}$. Let $\Vert\bm{D}\Vert_{F}=\delta_{D}$ and
$\Vert\bm{N}\Vert_{F}=\delta_{N}$, so $\delta_{D}^{2}+\delta_{N}^{2}=\delta^{2}$.
Hence,

\begin{align}
\left\Vert \bm{U}^{\circ3}-\bm{V}^{\circ3}\right\Vert _{F}^{2}\leq & \sum_{i,j\in[n]}\left(u_{i,j}^{3}-v_{i,j}^{3}\right)^{2}\nonumber \\
\leq & \sum_{i,j\in[n]}\left(u_{i,j}-v_{i,j}\right)^{2}\left(u_{i,j}^{2}+u_{i,j}v_{i,j}+v_{i,j}^{2}\right)^{2}\nonumber \\
\leq & \sum_{i\in[n]}\left(u_{i,i}-v_{i,i}\right)^{2}\left(u_{i,i}^{2}+u_{i,i}v_{i,i}+v_{i,i}^{2}\right)^{2}\nonumber \\
 & +\sum_{\substack{i\in[n]\\
j\neq i
}
}\left(u_{i,j}-v_{i,j}\right)^{2}\left(u_{i,j}^{2}+u_{i,j}v_{i,j}+v_{i,j}^{2}\right)^{2}\nonumber \\
\leq & 9\left(\delta_{D}^{2}+\epsilon^{4}\delta_{N}^{2}\right)\nonumber \\
\leq & 9\left(\frac{\delta_{D}^{2}}{\delta^{2}}+\epsilon^{4}\right)\delta^{2},
\end{align}

\noindent which implies that

\begin{equation}
\left\Vert \bm{U}^{\circ3}-\bm{V}^{\circ3}\right\Vert _{F}\leq3\sqrt{\frac{\delta_{D}^{2}}{\delta^{2}}+\epsilon^{4}}\delta.
\end{equation}

\noindent Since $\delta$ and $\epsilon$are known, we need to find
an upper bound of $\delta_{D}^{2}$, or equivalently, a lower bound
of $\delta_{N}^{2}$. We will accomplish this goal in two steps. In
the first step we will decouple each row of $\bm{U}$ and $\bm{V}$
into $n$ subproblems, and after that we will solve each subproblem.
In the second step we will derive the conclusion of the original problem
from the conclusion obtained by solving the subproblems.

\noindent $\textbf{1.\ Decouple and Solve Subproblems}$

\noindent Let $\delta_{i}=\left\Vert \bm{u}_{i,\ast}-\bm{v}_{i,\ast}\right\Vert _{2}$,
$\delta_{D,i}=\left\Vert \bm{d}_{i,\ast}\right\Vert _{2}$, $\delta_{N,i}=\left\Vert \bm{n}_{i,\ast}\right\Vert _{2}$,
$\forall i\in[n]$. If

\begin{align}
f(\delta,\epsilon) & =\max_{\substack{\Vert\bm{U}-\bm{V}\Vert_{F}=\delta\\
\Vert\bm{U}-\bm{P}\Vert_{F}\leq\epsilon\\
\Vert\bm{V}-\bm{P}\Vert_{F}\leq\epsilon
}
}\delta_{D}^{2},\\
g_{i}\left(\delta_{i},\epsilon\right) & =\max_{\substack{\Vert\bm{u}_{i,\ast}-\bm{v}_{i,\ast}\Vert_{2}=\delta_{i}\\
\Vert\bm{v}_{i,\ast}-\bm{p}_{i,\ast}\Vert_{2}\leq\epsilon\\
\Vert\bm{v}_{i,\ast}-\bm{p}_{i,\ast}\Vert_{2}\leq\epsilon
}
}\delta_{D,i}^{2},\ \forall i\in[n].
\end{align}

\noindent we know that

\begin{equation}
f(\delta,\epsilon)\leq\max_{\sum_{i\in[n]}\delta_{i}^{2}=\delta^{2}}\sum_{i\in[n]}g_{i}\left(\delta_{i},\epsilon\right).
\end{equation}

\noindent Therefore, we can decouple each row of $\bm{U}$ and $\bm{V}$
into $n$ subproblems, solving the $n$ easier subproblems first.
For the $i^{\text{th}}$ subproblem, since $\delta_{i}$ and $\epsilon$are
known, we need to find an upper bound of $\delta_{D,i}^{2}$, or equivalently,
a lower bound of $\delta_{N,i}^{2}$. Next we will consider the impact
of the choice of diagonal and non-diagonal entries of $\bm{U}$ and
$\bm{V}$ on $\delta_{N,i}^{2}$ in turn.

\noindent $\textbf{1.1.\ The Choice of Non-diagonal Entries}$

\noindent Since

\begin{align}
\delta_{N,i}^{2}= & \sum_{j\neq i}\left(u_{i,j}-v_{i,j}\right)^{2}\nonumber \\
= & \sum_{j\neq i}u_{i,j}^{2}+\sum_{j\neq i}v_{i,j}^{2}-2\sum_{j\neq i}u_{i,j}v_{i,j}\nonumber \\
= & 1-u_{i,i}^{2}+1-v_{i,i}^{2}-2\sum_{j\neq i}u_{i,j}v_{i,j},
\end{align}

\noindent we know that $\delta_{N,i}^{2}$ is minimized, no matter
what the values of $u_{i,i}$ and $v_{i,i}$, if

\begin{align}
u_{i,j} & =\begin{cases}
\sqrt{1-u_{i,i}^{2}}, & \text{if}\ j=(i+1)\mod n,\\
0, & \text{if}\ j\neq i\ \text{and}\ j\neq(i+1)\mod n,
\end{cases}
\end{align}

\begin{align}
v_{i,j} & =\begin{cases}
\sqrt{1-v_{i,i}^{2}}, & \text{if}\ j=(i+1)\mod n,\\
0, & \text{if}\ j\neq i\ \text{and}\ j\neq(i+1)\mod n.
\end{cases}
\end{align}

\noindent $\textbf{1.2.\ The Choice of Diagonal Entries}$

\noindent We will further consider the choice of diagonal entries
of $\bm{U}$ and $\bm{V}$ on the basis of the subsection 1.1. We
can assume that $p_{i,i}=1$ and $u_{i,i}\geq v_{i,i}$ without loss
of generality. Lemma \ref{lem:lem4} implies that

\begin{equation}
u_{i,i}\geq v_{i,i}\geq1-\frac{1}{2}\epsilon^{2}.
\end{equation}

\noindent Assume that $v_{i,i}=C+x$ and $u_{i,i}=C+\delta_{D,i}+x$
where $C=1-\frac{1}{2}\epsilon^{2}$, hence,

\begin{align}
\delta_{N,i}^{2}= & \left(\sqrt{1-(C+x)^{2}}-\sqrt{1-\left(C+\delta_{D,i}+x\right)^{2}}\right)^{2}\nonumber \\
= & 1-(C+x)^{2}+1-\left(C+\delta_{D,i}+x\right)^{2}\nonumber \\
 & -2\sqrt{1-(C+x)^{2}}\sqrt{1-\left(C+\delta_{D,i}+x\right)^{2}}.
\end{align}

\noindent Since

\begin{align}
\frac{\partial\delta_{N,i}^{2}}{\partial x}= & -2C-2x-2C-2\delta_{D,i}-2x\nonumber \\
 & +2\sqrt{\frac{1-(C+\delta_{D,i}+x)^{2}}{1-(C+x)^{2}}}(C+x)\nonumber \\
 & +2\sqrt{\frac{1-(C+x)^{2}}{1-(C+\delta_{D,i}+x)^{2}}}\left(C+\delta_{D,i}+x\right)\nonumber \\
\geq & 0,
\end{align}

\noindent we know that $x=0$, i.e., $v_{i,i}=C$ and $u_{i,i}=C+\delta_{D,i}$
can minimize $\delta_{N,i}^{2}$. 

\noindent $\textbf{2.\ Solve the Original Problem}$

\noindent In this section, we will unify $n$ subproblems again to
find a lower bound of $\delta_{N}^{2}$. Consider the optimization
problem (\ref{eq:eq18})

\begin{align}
\min_{\delta_{D,1},\ldots,\delta_{D,n}} & \quad\sum_{i\in[n]}\delta_{N,i}^{2}\nonumber \\
\mathrm{s.t.} & \quad\sum_{i\in[n]}\delta_{D,i}^{2}=\delta_{D}^{2},\nonumber \\
 & \quad\delta_{D,i}\geq0,\ \forall i\in[n],\label{eq:eq18}
\end{align}

\noindent where

\begin{align*}
\delta_{N,i}^{2}= & 1-C^{2}+1-\left(C+\delta_{D,i}\right)^{2}\\
 & -2\sqrt{1-C^{2}}\sqrt{1-\left(C+\delta_{D,i}\right)^{2}},\forall i\in[n].
\end{align*}

\noindent Substitute $\delta_{D,i}$ by $\sqrt{\Delta_{D,i}},\ \forall i\in[n]$,
(\ref{eq:eq18}) can be reformulated as

\begin{align}
\min_{\Delta_{D,1},\ldots,\Delta_{D,n}} & \quad\sum_{i\in[n]}\delta_{N,i}^{2}\nonumber \\
\mathrm{s.t.} & \quad\sum_{i\in[n]}\Delta_{D,i}=\delta_{D}^{2},\nonumber \\
 & \quad\Delta_{D,i}\geq0,\ \forall i\in[n],\label{eq:eq19}
\end{align}

\noindent where

\begin{align}
\delta_{N,i}^{2}= & 1-C^{2}+1-\left(C+\sqrt{\Delta_{D,i}}\right)^{2}\nonumber \\
 & -2\sqrt{1-C^{2}}\sqrt{1-\left(C+\sqrt{\Delta_{D,i}}\right)^{2}},\forall i\in[n].
\end{align}

\noindent Apply the Karush-Kuhn-Tucker conditions to (\ref{eq:eq19}),
we have

\begin{equation}
\begin{cases}
\frac{\sqrt{1-C^{2}}\left(C+\sqrt{\Delta_{D,i}}\right)}{\sqrt{\Delta_{D,i}}\sqrt{1-\left(C+\sqrt{\Delta_{D,i}}\right)^{2}}}-\frac{C+\sqrt{\Delta_{D,i}}}{\sqrt{\Delta_{D,i}}}+\lambda+\mu_{i}=0,\\
\mu_{i}\Delta_{D,i}=0,\ \forall i\in[n].
\end{cases}
\end{equation}

\noindent Let $\bm{S}=\left\{ i|\Delta_{D,i}>0\right\} $, we have

\begin{equation}
f_{1}\left(\Delta_{D,i}\right)=\frac{\sqrt{1-C^{2}}\left(C+\sqrt{\Delta_{D,i}}\right)}{\sqrt{\Delta_{D,i}}\sqrt{1-\left(C+\sqrt{\Delta_{D,i}}\right)^{2}}}-\frac{C+\sqrt{\Delta_{D,i}}}{\sqrt{\Delta_{D,i}}},
\end{equation}

\noindent for every $i\in\bm{S}$.

\noindent If $f_{1}\left(\Delta_{D,i}\right)$ is strictly monotonic
increasing in $\left(0,(1-C)^{2}\right)$, then $\Delta_{D,i}$ are
equal for every $i\in\bm{S}$. Since $f_{2}(x)=x^{2}$ is strictly
monotonic increasing in $(0,1-C)$, we only need to show that $f_{3}(x)=(f_{1}\circ f_{2})(x)$
is strictly monotonic increasing in $(0,1-C)$ if we want to show
that $f_{1}\left(\Delta_{D,i}\right)$ is strictly monotonic increasing
in $\left(0,(1-C)^{2}\right)$. The derivative of $f_{3}(x)$ is

\begin{align}
f_{3}'(x)= & \frac{\sqrt{1-C^{2}}x(C+x)^{2}+C\left(1-(C+x)^{2}\right)^{\frac{3}{2}}}{x^{2}\left(1-(C+x)^{2}\right)^{\frac{3}{2}}}\nonumber \\
 & -\frac{C\sqrt{1-C^{2}}\left(1-(C+x)^{2}\right)}{x^{2}\left(1-(C+x)^{2}\right)^{\frac{3}{2}}}.\label{eq:eq20}
\end{align}

\noindent Note that denominator of (\ref{eq:eq20}) is positive, i.e.,
$x^{2}\left(1-(C+x)^{2}\right)^{\frac{3}{2}}>0$, in $(0,1-C)$, so
we only need to show that the numerator of (\ref{eq:eq20}) is also
positive in $(0,1-C)$. Let

\begin{align}
f_{4}(x)= & \sqrt{1-C^{2}}x(C+x)^{2}+C\left(1-(C+x)^{2}\right)^{\frac{3}{2}}\nonumber \\
 & -C\sqrt{1-C^{2}}\left(1-(C+x)^{2}\right).
\end{align}

\noindent The derivative of $f_{4}(x)$ is

\begin{equation}
f_{4}'(x)=3(C+x)\left(\sqrt{1-C^{2}}(C+x)-C\sqrt{1-(C+x)^{2}}\right).
\end{equation}

\noindent Since $f_{4}(0)=0$, we only need to show that $f_{4}'(x)>0$
in $(0,1-C)$. Note that

\begin{align}
 & (C+x)^{2}>C^{2}\nonumber \\
\Rightarrow & (C+x)^{2}-C^{2}(C+x)^{2}>C^{2}-C^{2}(C+x)^{2}\nonumber \\
\Rightarrow & \left(1-C^{2}\right)(C+x)^{2}>C^{2}\left(1-(C+x)^{2}\right)\nonumber \\
\Rightarrow & \sqrt{1-C^{2}}(C+x)>C\sqrt{1-(C+x)^{2}},
\end{align}

\noindent we know that $f_{4}'(x)>0$ in $(0,1-C)$. Therefore, $\Delta_{D,i}$
are equal for every $i\in\bm{S}$ which implies that $\delta_{D,i}$
are equal for every $i\in\bm{S}$.

\noindent Now, we need to consider the cardinality of $\bm{S}$. Suppose
that $|\bm{S}|=m\in[1,n]$, then $\delta_{D,i}=\frac{\delta_{D}}{\sqrt{m}}$
and the objective of (\ref{eq:eq19}) is

\begin{align}
g_{1}(m)= & m\left(1-C^{2}+1-\left(C+\frac{\delta_{D}}{\sqrt{m}}\right)^{2}\right)\nonumber \\
 & -2m\sqrt{1-C^{2}}\sqrt{1-\left(C+\frac{\delta_{D}}{\sqrt{m}}\right)^{2}}.
\end{align}

\noindent Since $g_{2}(x)=\frac{\delta_{D}^{2}}{x^{2}}$ is strictly
monotonic decreasing in $(0,1-C)$, we only need to show that $g_{3}(x)=\frac{1}{2\delta_{D}^{2}}(g_{1}\circ g_{2})(x)$
is strictly monotonic increasing in $(0,1-C)$ if we want to show
that $g_{1}(m)$ is strictly monotonic decreasing in $\left(\frac{\delta_{D}^{2}}{(1-C)^{2}},+\infty\right)$.
The derivative of $g_{3}(x)$ is

\begin{align}
g_{3}'(x)= & \frac{\sqrt{1-C^{2}}x^{2}-2\sqrt{1-(C+x)^{2}}\left(1-C^{2}\right)}{x^{3}\sqrt{1-(C+x)^{2}}}\nonumber \\
 & +\frac{2\sqrt{1-C^{2}}\left(1-(C+x)^{2}\right)}{x^{3}\sqrt{1-(C+x)^{2}}}\nonumber \\
 & +\frac{\left(\sqrt{1-(C+x)^{2}}+\sqrt{1-C^{2}}\right)Cx}{x^{3}\sqrt{1-(C+x)^{2}}}.\label{eq:eq21}
\end{align}

\noindent Note that denominator of (\ref{eq:eq21}) is positive, i.e.,
$x^{3}\sqrt{1-(C+x)^{2}}>0$, in $(0,1-C)$, so we only need to show
that the numerator of (\ref{eq:eq21}) is also positive in $(0,1-C)$.
Let

\begin{align}
g_{4}(x)= & \sqrt{1-C^{2}}x^{2}-2\sqrt{1-(C+x)^{2}}\left(1-C^{2}\right)\nonumber \\
 & +2\sqrt{1-C^{2}}\left(1-(C+x)^{2}\right)\nonumber \\
 & +\left(\sqrt{1-(C+x)^{2}}+\sqrt{1-C^{2}}\right)Cx.
\end{align}

\noindent The derivative of $g_{4}(x)$ is
\begin{equation}
g_{4}'(x)=(3C+2x)\frac{1-Cx-C^{2}-\sqrt{1-C^{2}}\sqrt{1-(C+x)^{2}}}{\sqrt{1-(C+x)^{2}}}.
\end{equation}

\noindent Since $g_{4}(0)=0$, we only need to show that $g_{4}'(x)>0$
in $(0,1-C)$. Note that

\begin{align}
 & C^{4}+2C^{3}x+C^{2}x^{2}-2C^{2}-2Cx-x^{2}+1\nonumber \\
 & <C^{4}+2C^{3}x+C^{2}x^{2}-2C^{2}-2Cx+1\nonumber \\
\Rightarrow & \left(1-C^{2}\right)\left(1-(C+x)^{2}\right)<\left(1-Cx+C^{2}\right)^{2}\nonumber \\
\Rightarrow & \sqrt{1-C^{2}}\sqrt{1-(C+x)^{2}}<1-Cx+C^{2},
\end{align}

\noindent we know that $g_{4}'(x)>0$ in $(0,1-C)$ and $|\bm{S}|=n$.
Hence, a lower bound of $\delta_{N}^{2}$ is

\begin{align}
\delta_{N}^{2}\geq & n\left(1-C^{2}+1-\left(C+\frac{\delta_{D}}{\sqrt{n}}\right)^{2}\right)\nonumber \\
 & -2n\sqrt{1-C^{2}}\sqrt{1-\left(C+\frac{\delta_{D}}{\sqrt{n}}\right)^{2}}.
\end{align}

\noindent Let $C_{1}=\sqrt{1-C^{2}},C_{2}=\sqrt{1-\left(C+\frac{\delta_{D}}{\sqrt{n}}\right)^{2}}$,
we have

\begin{align}
\delta_{N}^{2}\geq & n\left(C_{1}+C_{2}\right)^{2}\nonumber \\
= & n\left(\frac{C_{1}^{2}-C_{2}^{2}}{C_{1}-C_{2}}\right)^{2}\nonumber \\
\geq & \frac{n\left(2C\frac{\delta_{D}}{\sqrt{n}}+\frac{\delta_{D}^{2}}{n}\right)^{2}}{4\left(1-C^{2}\right)}\nonumber \\
\geq & \frac{C^{2}}{1-C^{2}}\delta_{D}^{2},
\end{align}

\noindent which implies that

\begin{align}
\frac{\delta_{D}^{2}}{\delta^{2}}\leq & \frac{\delta_{D}^{2}}{\delta_{D}^{2}+\frac{C^{2}}{1-C^{2}}\delta_{D}^{2}}\nonumber \\
= & 1-\left(1-\frac{\epsilon^{2}}{2}\right)^{2}\nonumber \\
\leq & 2\epsilon^{2}-\epsilon^{4}.
\end{align}

\noindent Therefore, we can conclude that

\begin{align}
\left\Vert \bm{U}^{\circ3}-\bm{V}^{\circ3}\right\Vert _{F} & \leq3\sqrt{\frac{\delta_{D}^{2}}{\delta^{2}}+\epsilon^{4}}\delta\nonumber \\
 & \leq3\sqrt{2}\epsilon\Vert\bm{U}-\bm{V}\Vert_{F}.
\end{align}
\end{IEEEproof}
\begin{lem}
\label{lem:lem8} Let $\bm{U}$,$\bm{U}_{1}$,$\ldots$,$\bm{U}_{N}\in\mathrm{O}(n)$
such that $\Vert\bm{U}-\bm{U}_{i}\Vert_{F}\leq\delta$, $\forall i\in[N]$,
and let $\bm{X}=\left[\bm{X}_{1},\ldots,\bm{X}_{N}\right]\in\mathbb{R}^{n\times p}$,
$x_{i,j}\sim_{i.i.d.}\mathrm{BG}(\theta)$ where $\bm{X}_{i}\in\mathbb{R}^{n\times p_{i}}$,
$\forall i\in[N]$. If there exists a signed permutation matrix $\bm{P}$,
such that $\Vert\bm{U}-\bm{P}\Vert_{F}\leq\epsilon\leq1$ and $\Vert\bm{U}_{i}-\bm{P}\Vert_{F}\leq\epsilon\leq1$,
$\forall i\in[N]$, then

\begin{align}
 & \left\Vert \mathbb{E}\left[\left(\bm{UX}\right)^{\circ3}\bm{X}^{\mathsf{T}}\right]-\mathbb{E}\left[\sum_{i\in[N]}\left(\bm{U}_{i}\bm{X}_{i}\right)^{\circ3}\bm{X}_{i}^{\mathsf{T}}\right]\right\Vert _{F}\nonumber \\
\leq & \left(9\sqrt{2}p\theta(1-\theta)\epsilon+3p\theta^{2}\right)\delta.
\end{align}
\end{lem}
\begin{IEEEproof}
\noindent (98) in \cite{zhai2020complete} and Lemma \ref{lem:lem8}
implies that

\begin{align}
 & \left\Vert \mathbb{E}\left[\left(\bm{UX}\right)^{\circ3}\bm{X}^{\mathsf{T}}\right]-\mathbb{E}\left[\sum_{i\in[N]}\left(\bm{U}_{i}\bm{X}_{i}\right)^{\circ3}\bm{X}_{i}^{\mathsf{T}}\right]\right\Vert _{F}\nonumber \\
\leq & \sum_{i\in[N]}\left\Vert \mathbb{E}\left[\left(\bm{UX}_{i}\right)^{\circ3}\bm{X}_{i}^{\mathsf{T}}\right]-\mathbb{E}\left[\left(\bm{U}_{i}\bm{X}_{i}\right)^{\circ3}\bm{X}_{i}^{\mathsf{T}}\right]\right\Vert _{F}\nonumber \\
\leq & \sum_{i\in[N]}3p_{i}\theta(1-\theta)\left\Vert \bm{U}^{\circ3}-\bm{U}_{i}^{\circ3}\right\Vert _{F}+3p_{i}\theta^{2}\left\Vert \bm{U}-\bm{U}_{i}\right\Vert _{F}\nonumber \\
\leq & \sum_{i\in[N]}\left(9\sqrt{2}p_{i}\theta(1-\theta)\epsilon+3p_{i}\theta^{2}\right)\left\Vert \bm{U}-\bm{U}_{i}\right\Vert _{F}\nonumber \\
\leq & \left(9\sqrt{2}p\theta(1-\theta)\epsilon+3p\theta^{2}\right)\delta.
\end{align}
\end{IEEEproof}
\begin{lem}
\label{lem:lem9} Let $\bm{U}$,$\bm{U}_{1}$,$\ldots$,$\bm{U}_{N}\in\mathrm{O}(n)$
such that $\Vert\bm{U}-\bm{U}_{i}\Vert_{F}\leq\delta$, $\forall i\in[N]$,
and let $\bm{X}=\left[\bm{X}_{1},\ldots,\bm{X}_{N}\right]\in\mathbb{R}^{n\times p}$,
$x_{i,j}\sim_{i.i.d.}\mathrm{BG}(\theta)$ where $\bm{X}_{i}\in\mathbb{R}^{n\times p_{i}}$,
$\forall i\in[N]$. If $\bar{\bm{X}}$ is the truncation of $\bm{X}$
by bound $B$

\begin{equation}
\bar{x}_{i,j}=\begin{cases}
x_{i,j}, & \text{if}\ |x_{i,j}|\leq B,\\
0, & \text{otherwise},
\end{cases}
\end{equation}

\noindent then

\begin{equation}
\left\Vert \left(\bm{UX}\right)^{\circ3}\bm{X}^{\mathsf{T}}-\sum_{i\in[N]}\left(\bm{U}_{i}\bm{X}_{i}\right)^{\circ3}\bm{X}_{i}^{\mathsf{T}}\right\Vert _{F}\leq3n^{2}pB^{4}\delta.
\end{equation}
\end{lem}
\begin{IEEEproof}
(185), (186), and (187) in \cite{zhai2020complete} implies that

\begin{align}
 & \left\Vert \left(\bm{UX}\right)^{\circ3}\bm{X}^{\mathsf{T}}-\sum_{i\in[N]}\left(\bm{U}_{i}\bm{X}_{i}\right)^{\circ3}\bm{X}_{i}^{\mathsf{T}}\right\Vert _{F}\nonumber \\
\leq & \sum_{i\in[N]}\sum_{j\in\left[p_{i}\right]}\bigg\Vert\left(\bm{U}\left\{ \bar{\bm{X}}_{i}\right\} _{*,j}\right)^{\circ3}\left\{ \bar{\bm{X}}_{i}\right\} _{*,j}^{\mathsf{T}}\nonumber \\
 & -\left(\bm{U}_{i}\left\{ \bar{\bm{X}}_{i}\right\} _{*,j}\right)^{\circ3}\left\{ \bar{\bm{X}}_{i}\right\} _{*,j}^{\mathsf{T}}\bigg\Vert_{F}\nonumber \\
\leq & 3\sum_{i\in[N]}\sum_{j\in\left[p_{i}\right]}\left\Vert \left\{ \bar{\bm{X}}_{i}\right\} _{*,j}\right\Vert _{F}^{4}\left\Vert \bm{U}-\bm{U}_{i}\right\Vert _{F}\nonumber \\
\leq & 3n^{2}pB^{4}\delta.
\end{align}
\end{IEEEproof}
\begin{lem}
\label{lem:lem10} Let $\bm{U}_{1}$,$\ldots$,$\bm{U}_{N}\in\mathrm{O}(n)$
such that $\left\Vert \bm{U}_{i}-\bm{U}_{j}\right\Vert _{F}\leq\delta_{c}$,
$\forall i$,$j\in[N]$ and let $\bm{X}=\left[\bm{X}_{1},\ldots,\bm{X}_{N}\right]\in\mathbb{R}^{n\times p}$,
$x_{i,j}\sim_{i.i.d.}\mathrm{BG}(\theta)$ where $\bm{X}_{i}\in\mathbb{R}^{n\times p_{i}}$,
$\forall i\in[N]$. If $\xi\geq8n(\ln p)^{4}\delta_{c}$, then

\begin{align}
 & \mathbb{P}\left(\left\Vert \sum_{i\in[N]}\left(\bm{U}_{i}\bm{X}_{i}\right)^{\circ3}\bm{X}_{i}^{\mathsf{T}}-\mathbb{E}\left[\left(\bm{U}_{i}\bm{X}_{i}\right)^{\circ3}\bm{X}_{i}^{\mathsf{T}}\right]\right\Vert _{F}\geq np\xi\right)\nonumber \\
\leq & 2n^{2}\exp\left(\frac{-3p\xi^{2}}{c_{1}\theta+8n^{\frac{3}{2}}(\ln p)^{4}\xi}\right)+2np\theta\exp\left(\frac{-(\ln p)^{2}}{2}\right).
\end{align}

\noindent for a consant $c_{1}>1.7\times10^{4}$. 
\end{lem}
\begin{IEEEproof}
Let $\bar{\bm{X}}$ be the truncation of $\bm{X}$ by bound $\ln p$

\begin{equation}
\bar{x}_{i,j}=\begin{cases}
x_{i,j}, & \text{if}\ |x_{i,j}|\leq\ln p,\\
0 & \text{otherwise}.
\end{cases}
\end{equation}

\noindent Let $\bm{U}=\bm{U}_{1}$. Since $\left\Vert \bm{U}_{i}-\bm{U}_{j}\right\Vert _{F}\leq\delta_{c}$,
$\forall i$,$j\in[N]$, we know that

\begin{equation}
\left\Vert \bm{U}-\bm{U}_{i}\right\Vert _{F}\leq\delta_{c},\ \forall i\in[N].
\end{equation}

\noindent Lemma 34 in \cite{zhai2020complete} implies that

\begin{align}
 & \mathbb{P}\left(\left\Vert \sum_{i\in[N]}\left(\bm{U}_{i}\bm{X}_{i}\right)^{\circ3}\bm{X}_{i}^{\mathsf{T}}-\mathbb{E}\left[\left(\bm{U}_{i}\bm{X}_{i}\right)^{\circ3}\bm{X}_{i}^{\mathsf{T}}\right]\right\Vert _{F}\geq np\xi\right)\nonumber \\
\leq & \mathbb{P}\left(\left\Vert \sum_{i\in[N]}\left(\bm{U}_{i}\bar{\bm{X}}_{i}\right)^{\circ3}\bar{\bm{X}}_{i}^{\mathsf{T}}-\mathbb{E}\left[\left(\bm{U}_{i}\bm{X}_{i}\right)^{\circ3}\bm{X}_{i}^{\mathsf{T}}\right]\right\Vert _{F}\geq np\xi\right)\nonumber \\
 & +\mathbb{P}(\bm{X}\neq\bar{\bm{X}})\nonumber \\
\leq & \underbrace{\mathbb{P}\left(\left\Vert \sum_{i\in[N]}\left(\bm{U}_{i}\bar{\bm{X}}_{i}\right)^{\circ3}\bar{\bm{X}}_{i}^{\mathsf{T}}-\mathbb{E}\left[\left(\bm{U}_{i}\bm{X}_{i}\right)^{\circ3}\bm{X}_{i}^{\mathsf{T}}\right]\right\Vert _{F}\geq np\xi\right)}_{\Gamma}\nonumber \\
 & +2np\theta\left(\frac{-(\ln p)^{2}}{2}\right).\label{eq:eq22}
\end{align}

\noindent Now we only need to obtain an upper bound of $\Gamma$.
By Lemma \ref{lem:lem8}, Lemma \ref{lem:lem9}, we have

\begin{align}
\Gamma\leq & \mathbb{P}\vast(\left\Vert \sum_{i\in[N]}\left(\bm{U}_{i}\bar{\bm{X}}_{i}\right)^{\circ3}\bar{\bm{X}}_{i}^{\mathsf{T}}-\left(\bm{U}\bar{\bm{X}}\right)^{\circ3}\bar{\bm{X}}^{\mathsf{T}}\right\Vert _{F}\nonumber \\
 & +\left\Vert \mathbb{E}\left[\left(\bm{UX}\right)^{\circ3}\bm{X}^{\mathsf{T}}\right]-\mathbb{E}\left[\sum_{i\in[N]}\left(\bm{U}_{i}\bm{X}_{i}\right)^{\circ3}\bm{X}_{i}^{\mathsf{T}}\right]\right\Vert _{F}\nonumber \\
 & +\left\Vert \left(\bm{U}\bar{\bm{X}}\right)^{\circ3}\bar{\bm{X}}^{\mathsf{T}}-\mathbb{E}\left[\left(\bm{UX}\right)^{\circ3}\bm{X}^{\mathsf{T}}\right]\right\Vert _{F}\geq np\xi\vast)\nonumber \\
\leq & \mathbb{P}\bigg(\left\Vert \left(\bm{U}\bar{\bm{X}}\right)^{\circ3}\bar{\bm{X}}^{\mathsf{T}}-\mathbb{E}\left[\left(\bm{UX}\right)^{\circ3}\bm{X}^{\mathsf{T}}\right]\right\Vert _{F}\nonumber \\
 & +\left(3n^{2}p(\ln p)^{4}+9p\sqrt{n}\theta(1-\theta)+3p\theta^{2}\right)\delta_{c}\geq np\xi\bigg)\nonumber \\
\leq & \mathbb{P}\bigg(\left\Vert \left(\bm{U}\bar{\bm{X}}\right)^{\circ3}\bar{\bm{X}}^{\mathsf{T}}-\mathbb{E}\left[\left(\bm{UX}\right)^{\circ3}\bm{X}^{\mathsf{T}}\right]\right\Vert _{F}\nonumber \\
 & +4n^{2}p(\ln p)^{4}\delta_{c}\geq np\xi\bigg).
\end{align}

\noindent Since $\xi\geq8n(\ln p)^{4}\delta_{c}$, we have
\begin{align}
\Gamma\leq & \mathbb{P}\left(\left\Vert \left(\bm{U}\bar{\bm{X}}\right)^{\circ3}\bar{\bm{X}}^{\mathsf{T}}-\mathbb{E}\left[\left(\bm{UX}\right)^{\circ3}\bm{X}^{\mathsf{T}}\right]\right\Vert _{F}\geq\frac{np\xi}{2}\right).
\end{align}

\noindent By (232), (233), and (239) in \cite{zhai2020complete},
we have

\noindent 
\begin{align}
\Gamma\leq & \mathbb{P}\left(\left\Vert \left(\bm{U}\bar{\bm{X}}\right)^{\circ3}\bar{\bm{X}}^{\mathsf{T}}-\mathbb{E}\left[\left(\bm{U\bar{X}}\right)^{\circ3}\bm{\bar{X}}^{\mathsf{T}}\right]\right\Vert _{F}\geq\frac{np\xi}{4}\right)\nonumber \\
\leq & 2n^{2}\exp\left(\frac{-3p\xi^{2}}{c_{1}\theta+8n^{\frac{3}{2}}(\ln p)^{4}\xi}\right).\label{eq:eq23}
\end{align}

\noindent for a consant $c_{1}>1.7\times10^{4}$. 

\noindent Combine (\ref{eq:eq22}) and (\ref{eq:eq23}), we have
\begin{align}
 & \mathbb{P}\left(\left\Vert \sum_{i\in[N]}\left(\bm{U}_{i}\bm{X}_{i}\right)^{\circ3}\bm{X}_{i}^{\mathsf{T}}-\mathbb{E}\left[\left(\bm{U}_{i}\bm{X}_{i}\right)^{\circ3}\bm{X}_{i}^{\mathsf{T}}\right]\right\Vert _{F}\geq np\xi\right)\nonumber \\
\leq & 2n^{2}\exp\left(\frac{-3p\xi^{2}}{c_{1}\theta+8n^{\frac{3}{2}}(\ln p)^{4}\xi}\right)+2np\theta\exp\left(-\frac{(\ln p)^{2}}{2}\right).
\end{align}

\noindent for a consant $c_{1}>1.7\times10^{4}$. 
\end{IEEEproof}
\begin{lem}
\label{lem:lem11} Let $\bm{U}$,$\bm{U}_{1}$,$\ldots$,$\bm{U}_{N}\in\mathrm{O}(n)$
such that $\Vert\bm{U}-\bm{U}_{i}\Vert_{F}\leq\delta$, $\left\Vert \bm{U}_{i}-\bm{U}_{j}\right\Vert _{F}\leq\delta_{c}$,
$\forall i$,$j\in[N]$ and let $\bm{X}=\left[\bm{X}_{1},\ldots,\bm{X}_{N}\right]\in\mathbb{R}^{n\times p}$,
$x_{i,j}\sim_{i.i.d.}\mathrm{BG}(\theta)$ where $\bm{X}_{i}\in\mathbb{R}^{n\times p_{i}}$,
$\forall i\in[N]$. If there exists a signed permutation matrix $\bm{P}\in\mathbb{R}^{n\times n}$,
such that $\Vert\bm{U}-\bm{P}\Vert_{F}=\epsilon\leq1$, $\Vert\bm{U}_{i}-\bm{P}\Vert_{F}\leq\epsilon\leq1$,
$\forall i\in[N]$, and $\xi\geq16n(\ln p)^{4}\delta_{c}$, then\textup{ }

\begin{align}
 & \mathbb{P}\vast(\left\Vert \left(\bm{UX}\right)^{\circ3}\bm{X}^{\mathsf{T}}-\sum_{i\in[N]}\left(\bm{U}_{i}\bm{X}_{i}\right)^{\circ3}\bm{X}_{i}^{\mathsf{T}}\right\Vert _{F}\nonumber \\
 & \geq\left(9\sqrt{2}p\theta(1-\theta)\epsilon+3p\theta^{2}\right)\delta+np\xi\vast)\nonumber \\
\leq & 4n^{2}\exp\left(\frac{-3p\xi^{2}}{c_{2}\theta+16n^{\frac{3}{2}}(\ln p)^{4}\xi}\right)+4np\theta\exp\left(\frac{-(\ln p)^{2}}{2}\right),
\end{align}

\noindent for a constant $c_{2}>6.8\times10^{4}$. 
\end{lem}
\begin{IEEEproof}
The triangle inequality, Lemma \ref{lem:lem8}, and Lemma \ref{lem:lem10}
implies that
\begin{align}
 & \mathbb{P}\vast(\left\Vert \left(\boldsymbol{UX}\right)^{\circ3}\boldsymbol{X}^{\mathsf{T}}-\sum_{i\in[N]}\left(\boldsymbol{U}_{i}\boldsymbol{X}_{i}\right)^{\circ3}\boldsymbol{X}_{i}^{\mathsf{T}}\right\Vert _{F}\nonumber \\
 & <\left(9\sqrt{2}p\theta(1-\theta)\epsilon+3p\theta^{2}\right)\delta+np\xi\vast)\nonumber \\
\geq & \mathbb{P}\vast(\left\Vert \left(\bm{UX}\right)^{\circ3}\bm{X}^{\mathsf{T}}-\mathbb{E}\left[\left(\bm{UX}\right)^{\circ3}\bm{X}^{\mathsf{T}}\right]\right\Vert _{F}\nonumber \\
 & +\left\Vert \mathbb{E}\left[\left(\bm{UX}\right)^{\circ3}\bm{X}^{\mathsf{T}}\right]-\mathbb{E}\left[\sum_{i\in[N]}\left(\bm{U}_{i}\bm{X}_{i}\right)^{\circ3}\bm{X}_{i}^{\mathsf{T}}\right]\right\Vert _{F}\nonumber \\
 & +\left\Vert \mathbb{E}\left[\sum_{i\in[N]}\left(\bm{U}_{i}\bm{X}_{i}\right)^{\circ3}\bm{X}_{i}^{\mathsf{T}}\right]-\sum_{i\in[N]}\left(\bm{U}_{i}\bm{X}_{i}\right)^{\circ3}\bm{X}_{i}^{\mathsf{T}}\right\Vert _{F}\nonumber \\
 & <\left(9\sqrt{2}p\theta(1-\theta)\epsilon+3p\theta^{2}\right)\delta+np\xi\vast)\nonumber \\
\geq & \mathbb{P}\vast(\left\Vert \mathbb{E}\left[\sum_{i\in[N]}\left(\bm{U}_{i}\bm{X}_{i}\right)^{\circ3}\bm{X}_{i}^{\mathsf{T}}\right]-\sum_{i\in[N]}\left(\bm{U}_{i}\bm{X}_{i}\right)^{\circ3}\bm{X}_{i}^{\mathsf{T}}\right\Vert _{F}\nonumber \\
 & +\left\Vert \left(\bm{UX}\right)^{\circ3}\bm{X}^{\mathsf{T}}-\mathbb{E}\left[\left(\bm{UX}\right)^{\circ3}\bm{X}^{\mathsf{T}}\right]\right\Vert _{F}<np\xi\vast)\nonumber \\
\geq & 1-\mathbb{P}\left(\left\Vert \left(\bm{UX}\right)^{\circ3}\bm{X}^{\mathsf{T}}-\mathbb{E}\left[\left(\bm{UX}\right)^{\circ3}\bm{X}^{\mathsf{T}}\right]\right\Vert _{F}\geq\frac{np\xi}{2}\right)\nonumber \\
 & -\mathbb{P}\vast(\left\Vert \mathbb{E}\left[\sum_{i\in[N]}\left(\bm{U}_{i}\bm{X}_{i}\right)^{\circ3}\bm{X}_{i}^{\mathsf{T}}\right]-\sum_{i\in[N]}\left(\bm{U}_{i}\bm{X}_{i}\right)^{\circ3}\bm{X}_{i}^{\mathsf{T}}\right\Vert _{F}\nonumber \\
 & \geq\frac{np\xi}{2}\vast)\nonumber \\
\geq & 1-4n^{2}\exp\left(\frac{-3p\xi^{2}}{c_{2}\theta+16n^{\frac{3}{2}}(\ln p)^{4}\xi}\right)-4np\theta\exp\left(\frac{-(\ln p)^{2}}{2}\right),
\end{align}

\noindent for a constant $c_{2}>6.8\times10^{4}$. 
\end{IEEEproof}
\begin{lem}
\label{lem:lem12} Let $\bm{U}$,$\bm{U}_{1}$,$\ldots$,$\bm{U}_{N}\in\mathrm{O}(n)$
such that $\Vert\bm{U}-\bm{U}_{i}\Vert_{F}\leq\delta$, $\left\Vert \bm{U}_{i}-\bm{U}_{j}\right\Vert _{F}\leq\delta_{c}$,
$\forall i$,$j\in[N]$ and let $\bm{X}=\left[\bm{X}_{1},\ldots,\bm{X}_{N}\right]\in\mathbb{R}^{n\times p}$,
$x_{i,j}\sim_{i.i.d.}\mathrm{BG}(\theta)$ where $\bm{X}_{i}\in\mathbb{R}^{n\times p_{i}}$,
$\forall i\in[N]$. If there exists a signed permutation matrix $\bm{P}\in\mathbb{R}^{n\times n}$,
such that $\Vert\bm{U}-\bm{P}\Vert_{F}=\epsilon\leq1$, $\Vert\bm{U}_{i}-\bm{P}\Vert_{F}\leq\epsilon\leq1$,
$\forall i\in[N]$, then\textup{ }

\begin{align}
\sigma_{n}\left(\mathbb{E}\left[\sum_{i\in[N]}\left(\bm{U}_{i}\bm{X}_{i}\right)^{\circ3}\bm{X}_{i}^{\mathsf{T}}\right]\right)\nonumber \\
\geq3p\theta(1-\theta)(1-2n\epsilon)+3p\theta^{2}- & \left(9\sqrt{2}p\theta(1-\theta)\epsilon+3p\theta^{2}\right)\delta.\label{eq:eq24}
\end{align}
\end{lem}
\begin{IEEEproof}
(\ref{eq:eq24}) can obtained through Lemma \ref{lem:lem6}, Lemma
\ref{lem:lem8}, and Theorem 7.3.5 in \cite{horn2012matrix}.
\end{IEEEproof}
\begin{lem}
\label{lem:lem13} Let $\bm{U}$,$\bm{U}_{1}$,$\ldots$,$\bm{U}_{N}\in\mathrm{O}(n)$
such that $\Vert\bm{U}-\bm{U}_{i}\Vert_{F}\leq\delta$, $\left\Vert \bm{U}_{i}-\bm{U}_{j}\right\Vert _{F}\leq\delta_{c}$,
$\forall i$,$j\in[N]$ and let $\bm{X}=\left[\bm{X}_{1},\ldots,\bm{X}_{N}\right]\in\mathbb{R}^{n\times p}$,
$x_{i,j}\sim_{i.i.d.}\mathrm{BG}(\theta)$ where $\bm{X}_{i}\in\mathbb{R}^{n\times p_{i}}$,
$\forall i\in[N]$. If there exists a signed permutation matrix $\bm{P}\in\mathbb{R}^{n\times n}$,
such that $\Vert\bm{U}-\bm{P}\Vert_{F}=\epsilon\leq1$, $\Vert\bm{U}_{i}-\bm{P}\Vert_{F}\leq\epsilon\leq1$,
$\forall i\in[N]$, and $\xi\geq16n(\ln p)^{4}\delta_{c}$, then\textup{ }

\begin{align}
 & \mathbb{P}\vast(\sigma_{n}\left(\left(\boldsymbol{UX}\right)^{\circ3}\boldsymbol{X}^{\mathsf{T}}\right)+\sigma_{n}\left(\sum_{i\in[N]}\left(\bm{U}_{i}\bm{X}_{i}\right)^{\circ3}\bm{X}_{i}^{\mathsf{T}}\right)\leq6p\theta^{2}\nonumber \\
 & +6p\theta(1-\theta)(1-2n\epsilon)-\left(9\sqrt{2}p\theta(1-\theta)\epsilon+3p\theta^{2}\right)\delta-np\xi\vast)\nonumber \\
\leq & 4n^{2}\exp\left(\frac{-3p\xi^{2}}{c_{2}\theta+16n^{\frac{3}{2}}(\ln p)^{4}\xi}\right)+4np\theta\exp\left(\frac{-(\ln p)^{2}}{2}\right),
\end{align}

\noindent for a constant $c_{2}>6.8\times10^{4}$. 
\end{lem}
\begin{IEEEproof}
Lemma \ref{lem:lem6} and Lemma \ref{lem:lem12} implies that

\begin{align}
 & \sigma_{n}\left(\mathbb{E}\left[(\bm{UX})^{\circ3}\bm{X}^{\mathsf{T}}\right]\right)+\sigma_{n}\left(\mathbb{E}\left[\sum_{i\in[N]}\left(\bm{U}_{i}\bm{X}_{i}\right)^{\circ3}\bm{X}_{i}^{\mathsf{T}}\right]\right)\nonumber \\
\geq & 6p\theta(1-\theta)(1-2n\epsilon)+6p\theta^{2}-\left(9\sqrt{2}p\theta(1-\theta)\epsilon+3p\theta^{2}\right)\delta.
\end{align}

\noindent By Lemma \ref{lem:lem10}, we can further know that

\begin{align}
 & \mathbb{P}\vast(\sigma_{n}\left(\left(\boldsymbol{UX}\right)^{\circ3}\boldsymbol{X}^{\mathsf{T}}\right)+\sigma_{n}\left(\sum_{i\in[N]}\left(\bm{U}_{i}\bm{X}_{i}\right)^{\circ3}\bm{X}_{i}^{\mathsf{T}}\right)>6p\theta^{2}\nonumber \\
 & +6p\theta(1-\theta)(1-2n\epsilon)-\left(9\sqrt{2}p\theta(1-\theta)\epsilon+3p\theta^{2}\right)\delta-np\xi\vast)\nonumber \\
\geq & 1-\mathbb{P}\left(\left\Vert \left(\bm{UX}\right)^{\circ3}\bm{X}^{\mathsf{T}}-\mathbb{E}\left[\left(\bm{UX}\right)^{\circ3}\bm{X}^{\mathsf{T}}\right]\right\Vert _{F}\geq\frac{np\xi}{2}\right)\nonumber \\
 & -\mathbb{P}\vast(\left\Vert \mathbb{E}\left[\sum_{i\in[N]}\left(\bm{U}_{i}\bm{X}_{i}\right)^{\circ3}\bm{X}_{i}^{\mathsf{T}}\right]-\sum_{i\in[N]}\left(\bm{U}_{i}\bm{X}_{i}\right)^{\circ3}\bm{X}_{i}^{\mathsf{T}}\right\Vert _{F}\nonumber \\
 & \geq\frac{np\xi}{2}\vast)\nonumber \\
\geq & 1-4n^{2}\exp\left(\frac{-3p\xi^{2}}{c_{2}\theta+16n^{\frac{3}{2}}(\ln p)^{4}\xi}\right)-4np\theta\exp\left(\frac{-(\ln p)^{2}}{2}\right),
\end{align}

\noindent for a constant $c_{2}>6.8\times10^{4}$. 
\end{IEEEproof}

\section{Proof of Theorem \ref{thm:thm2} \label{sec:prf2}}
\begin{IEEEproof}
Lemma \ref{lem:lem6}, Lemma \ref{lem:lem8}, and Lemma \ref{lem:lem12}
implies that

\begin{align}
 & \frac{2\left\Vert \mathbb{E}\left[\bm{B}^{(t)}\right]-\mathbb{E}\left[\sum_{i\in[N]}\bm{B}_{i}^{(t)}\right]\right\Vert _{F}}{\sigma_{n}\left(\mathbb{E}\left[\bm{B}^{(t)}\right]\right)+\sigma_{n}\left(\mathbb{E}\left[\sum_{i\in[N]}\bm{B}_{i}^{(t)}\right]\right)}\nonumber \\
\leq & \frac{2\left(9\sqrt{2}p\theta(1-\theta)\epsilon+3p\theta^{2}\right)\delta^{(t)}}{6p\theta(1-\theta)(1-2n\epsilon)+6p\theta^{2}-\left(9\sqrt{2}p\theta(1-\theta)\epsilon+3p\theta^{2}\right)\delta^{(t)}}.
\end{align}

\noindent Hence, we only need to show that

\begin{align}
 & 6\alpha p\theta(1-\theta)(1-2n\epsilon)+6\alpha p\theta^{2}\nonumber \\
 & -\left(9\sqrt{2}\alpha p\theta(1-\theta)\epsilon+3\alpha p\theta^{2}\right)\delta^{(t)}\nonumber \\
> & 18\sqrt{2}p\theta(1-\theta)\epsilon+6p\theta^{2}.
\end{align}

\noindent Since $\delta^{(t)}\leq\max_{i\in[N]}\left\Vert \bm{U}_{i}^{(t-1)}-\bm{P}\right\Vert _{F}+\left\Vert \bm{U}^{(t-1)}-\bm{P}\right\Vert _{F}\leq2\epsilon$
and $\epsilon\leq$1, we only need to show that
\begin{align}
 & 6\alpha p\theta(1-\theta)(1-2n\epsilon)+6\alpha p\theta^{2}\nonumber \\
 & -\left(18\sqrt{2}\alpha p\theta(1-\theta)+6\alpha p\theta^{2}\right)\epsilon\nonumber \\
> & 18\sqrt{2}p\theta(1-\theta)\epsilon+6p\theta^{2}.
\end{align}

\noindent Since $\epsilon\in\left[0,\frac{\alpha-\theta}{2\alpha n(1-\theta)+3\sqrt{2}(1-\alpha)(1+\theta)+\alpha\theta}\right)$,
we can know that the inequality (\ref{eq:eq25}) holds.
\end{IEEEproof}

\section{Proof of Theorem \ref{thm:thm3} \label{sec:prf3}}
\begin{IEEEproof}
Let $\left\Vert \bm{B}^{(t)}-\sum_{i\in[N]}\bm{B}_{i}^{(t)}\right\Vert _{F}=\left(9\sqrt{2}p\theta(1-\theta)\epsilon+3p\theta^{2}\right)\delta^{(t)}+np\xi_{1}$,
$\sigma_{n}\left(\bm{B}^{(t)}\right)+\sigma_{n}\left(\sum_{i\in[N]}\bm{B}_{i}^{(t)}\right)=6p\theta(1-\theta)(1-2n\epsilon)+6p\theta^{2}-\left(9\sqrt{2}p\theta(1-\theta)\epsilon+3p\theta^{2}\right)\delta^{(t)}-np\xi_{2}$. 

\noindent Theorem \ref{thm:thm2} implies that 

\begin{align}
 & \mathbb{P}\left(\frac{2\left\Vert \bm{B}^{(t)}-\sum_{i\in[N]}\bm{B}_{i}^{(t)}\right\Vert _{F}}{\sigma_{n}\left(\bm{B}^{(t)}\right)+\sigma_{n}\left(\sum_{i\in[N]}\bm{B}_{i}^{(t)}\right)}<\alpha\delta^{(t)}\right)\nonumber \\
\geq & \mathbb{P}\left(\frac{2np\xi_{1}}{\delta^{(t)}}+np\xi_{2}<pC\right)\nonumber \\
\geq & \mathbb{P}\left(\frac{2np\xi_{1}}{\delta^{(t)}}<\frac{2}{\delta^{(t)}+2}pC,np\xi_{2}<\frac{\delta^{(t)}}{\delta^{(t)}+2}pC\right)\nonumber \\
\geq & 1-\mathbb{P}\left(np\xi_{1}\geq\frac{\delta^{(t)}}{\delta^{(t)}+2}pC\right)-\mathbb{P}\left(np\xi_{2}\geq\frac{\delta^{(t)}}{\delta^{(t)}+2}pC\right),
\end{align}

\noindent where $C=6\alpha\theta(1-\theta)(1-2n\epsilon)-6(1-\alpha)\theta^{2}-6\alpha\theta^{2}\epsilon-18\sqrt{2}(\alpha+1)\theta(1-\theta)\epsilon$.

\noindent By Lemma \ref{lem:lem11} and Lemma \ref{lem:lem13}, we
can know that the inequality (\ref{eq:eq26}) holds.
\end{IEEEproof}

\section{Proof of Theorem \ref{thm:thm4} \label{sec:prf4}}
\begin{IEEEproof}
Since

\begin{equation}
\delta_{c}^{(t+1)}\in\left[0,\alpha\delta^{(t)}-\frac{2\left\Vert \bm{B}^{(t)}-\sum_{i\in[N]}\bm{B}_{i}^{(t)}\right\Vert _{F}}{\sigma_{n}\left(\bm{B}^{(t)}\right)+\sigma_{n}\left(\sum_{i\in[N]}\bm{B}_{i}^{(t)}\right)}\right),
\end{equation}

\noindent we know that

\begin{align}
 & \mathbb{P}\left(\delta^{(t+1)}<\alpha\delta^{(t)}\right)\nonumber \\
\geq & \mathbb{P}\left(\delta_{c}^{(t+1)}+\delta_{a}^{(t+1)}<\alpha\delta^{(t)}\right)\nonumber \\
\geq & \mathbb{P}\left(\frac{2\left\Vert \bm{B}^{(t)}-\sum_{i\in[N]}\bm{B}_{i}^{(t)}\right\Vert _{F}}{\sigma_{n}\left(\bm{B}^{(t)}\right)+\sigma_{n}\left(\sum_{i\in[N]}\bm{B}_{i}^{(t)}\right)}<\alpha\delta^{(t)}\right)\nonumber \\
\geq & 1-8np\theta\exp\left(-\frac{(\ln p)^{2}}{2}\right)\nonumber \\
- & 8n^{2}\exp\left(\frac{-3p\left(\delta^{(t)}\right)^{2}C^{2}}{c_{2}\left(\delta^{(t)}+2\right)^{2}n^{2}\theta+16n^{\frac{5}{2}}(\ln p)^{4}\delta^{(t)}\left(\delta^{(t)}+2\right)C}\right),
\end{align}

\noindent where $C=6\alpha\theta(1-\theta)(1-2n\epsilon)-6(1-\alpha)\theta^{2}-6\alpha\theta^{2}\epsilon-18\sqrt{2}(\alpha+1)\theta(1-\theta)\epsilon$
and a constant $c_{2}>6.8\times10^{4}$.
\end{IEEEproof}

\section{Proof of Theorem \ref{thm:thm5} \label{sec:prf5}}
\begin{IEEEproof}
Lemma 6 in \cite{zhai2020complete}, (\ref{eq:eq28}), and (\ref{eq:eq29})
implies that

\begin{equation}
\left\Vert \bm{U}^{(t+1)}-\bm{P}\right\Vert _{F}\leq\epsilon-\delta^{(t)}\leq\epsilon-\delta^{(t+1)}.
\end{equation}

\noindent Since $\delta^{(t+1)}=\max_{i\in[N]}\left\Vert \bm{U}^{(t+1)}-\bm{U}_{i}^{(t+1)}\right\Vert _{F}$,
we know that

\begin{equation}
\max_{i\in[N]}\left\Vert \bm{U}_{i}-\bm{P}\right\Vert _{F}\leq\epsilon.
\end{equation}

\noindent Since $\left\Vert \bm{U}^{(t+1)}\right\Vert _{4}^{4}\geq\left\Vert \bm{U}^{(t)}\right\Vert _{4}^{4}$,
we know that

\begin{align}
\left\Vert \bm{U}^{(t+1)}\right\Vert _{4}^{4} & \text{\ensuremath{\geq}\ensuremath{\left\Vert \bm{U}^{(t)}\right\Vert _{4}^{4}}}\nonumber \\
 & \geq n-\frac{1}{2}\left(\epsilon-\delta^{(t)}\right)^{2}\nonumber \\
 & \geq n-\frac{1}{2}\left(\epsilon-\delta^{(t+1)}\right)^{2}.
\end{align}
\end{IEEEproof}
\bibliographystyle{IEEEtran}
\bibliography{DMSP}

\end{document}